\documentclass{article} 
\usepackage{iclr2019_conference,times}
\iclrfinalcopy

\usepackage{amsmath,amsfonts,bm}

\def\eqref#1{equation~\ref{#1}}

\def\1{\bm{1}}

\DeclareMathAlphabet{\mathsfit}{\encodingdefault}{\sfdefault}{m}{sl}
\SetMathAlphabet{\mathsfit}{bold}{\encodingdefault}{\sfdefault}{bx}{n}

\usepackage[utf8]{inputenc} 
\usepackage[T1]{fontenc}    
\usepackage{hyperref}
\usepackage{url}
\usepackage{booktabs}       
\usepackage{amsmath}
\usepackage{amssymb}
\usepackage{amsthm}
\usepackage{amsfonts}       
\usepackage{nicefrac}       
\usepackage{microtype}      
\usepackage[dvips]{graphicx}
\usepackage{color}
\usepackage{textcomp}
\usepackage{multirow}
\theoremstyle{plain}

\newtheorem{thm}{Theorem}
\newtheorem{prop}[thm]{Proposition}
\newtheorem{cor}[thm]{Corollary}

\newtheorem*{definition}{Definition}

\newcommand\blfootnote[1]{%
  \begingroup
  \renewcommand\thefootnote{}\footnote{#1}%
  \addtocounter{footnote}{-1}%
  \endgroup
}

\def\max{{\mbox{max}}}
\def\sup{{\mbox{sup}}}

\def\XX{{\sc MuLann}}
\def\DANN{{\sc Dann}}
\def\pheno{{\sc Cell}}
\def\office{{\sc Office}}
\def\digits{{\sc Digits}}

\def\sec{Sec.}
\title{Multi-Domain Adversarial Learning}

\author{
  Alice Schoenauer Sebag$^{1, \dagger}$\\
  \texttt{alice.schoenauer@polytechnique.org} \\
  \And
  Louise Heinrich$^1$,\\
  \texttt{louise.heinrich@ucsf.edu} \\
  \And
  Marc Schoenauer$^2$, \\
  \texttt{marc.schoenauer@inria.fr}\\
  \And
  Michele Sebag$^2$, \\
  \texttt{sebag@lri.fr}~~~~~~~~~~~~~~~~~~~~~~~~~~~~~~ \\
  \And
  Lani F. Wu$^1$,\\
  \texttt{lani.wu@ucsf.edu}~~~~~~~~~~~~~~~~~~~~~~~~~~~~~~ \\
  \And
  Steven J. Altschuler$^1$\\
  \texttt{steven.altschuler@ucsf.edu} \\
  \And
  $^1$ Department of Pharmaceutical Chemistry\\
  UCSF, San Francisco, CA 94158 \\
  \And
  $^2$ INRIA-CNRS-UPSud-UPSaclay \\
  TAU, U. Paris-Sud, 91405 Orsay\\
}
\begin{document}
\maketitle

\begin{abstract}
\blfootnote{\hspace{-0.05in}$\dagger$~Now at the French Ministry for the Economy and Finance, 75012 Paris.}Multi-domain learning (MDL) aims at obtaining a model with minimal average risk across multiple domains. Our empirical motivation is automated microscopy data, where cultured cells are imaged after being exposed to known and unknown chemical perturbations, and each dataset displays significant experimental bias. 
This paper presents a multi-domain adversarial learning approach, \XX, to leverage multiple datasets with overlapping but distinct class sets, in a semi-supervised setting.
Our contributions include: i) a bound on the average- and worst-domain risk in MDL, obtained using the $\mathcal{H}$-divergence; ii) a new loss to accommodate semi-supervised multi-domain learning and domain adaptation; iii) the experimental validation of the approach, improving on the state of the art on three standard image benchmarks, and a novel bioimage dataset, \pheno.\footnote{Code and data: \href{https://github.com/AltschulerWu-Lab/MuLANN}{github.com/AltschulerWu-Lab/MuLANN}}
\end{abstract}

\section{Introduction}

\label{introduction}
\vspace{-5pt}
Advances in technology have enabled large scale dataset generation by life sciences laboratories. These datasets contain information about overlapping but non-identical known and unknown experimental conditions. A challenge is how to best leverage information across multiple datasets on the same subject, and to make discoveries that could not have been obtained from any individual dataset alone. 

Transfer learning provides a formal framework for addressing this challenge, particularly crucial in cases where data acquisition is expensive and heavily impacted by experimental settings. One such field is automated microscopy, which can capture thousands of images of cultured cells after exposure to different experimental perturbations (e.g from chemical or genetic sources). A goal is to classify mechanisms by which perturbations affect cellular processes based on the similarity of cell images. In principle, it should be possible to tackle microscopy image classification as yet another visual object recognition task. However, two major challenges arise compared to mainstream visual object recognition problems~\citep{ILSVRC15}. First, biological images are heavily impacted by experimental choices, such as microscope settings and experimental reagents. Second, there is no standardized set of labeled perturbations, and datasets often contain labeled examples for a subset of possible classes only. This has limited microscopy image classification to single datasets and does not leverage the growing number of datasets collected by the life sciences community. These challenges make it desirable to learn models across many microscopy datasets, that achieve both good robustness w.r.t. experimental settings and good class coverage, all the while being robust to the fact that datasets contain samples from overlapping but distinct class sets. 

Multi-domain learning (MDL) aims to learn a model of minimal risk from datasets drawn from distinct underlying distributions~\citep{Dredze2010}, and is a particular case of transfer learning~\citep{5288526}. As such, it contrasts with the so-called domain adaptation (DA) problem~\citep{Scheffer,shai2010,ganin,5288526}. DA aims at learning a model with minimal risk on a distribution called "target" by leveraging other distributions called "sources". Notably, most DA methods assume that target classes are identical to source classes, or a subset thereof  in the case of partial DA~\citep{Cao_2018_partialDA, partialDA2}. 

The expected benefits of MDL, compared to training a separate model on each individual dataset, are two-fold. First, MDL leverages more (labeled and unlabeled) information, allowing better generalization while accommodating the specifics of each domain~\citep{Dredze2010, Xiao2016}. Thus, MDL models have a higher chance of \textit{ab initio} performing well on a new domain $-$ a problem referred to as domain generalization~\citep{krikamol} or zero-shot domain adaptation~\citep{Yang2015}. Second, MDL enables knowledge transfer between domains: in unsupervised and semi-supervised settings, concepts learned on one domain are applied to another, significantly reducing the need for labeled examples from the latter~\citep{5288526}.

Learning a single model from samples drawn from $n$ distributions raises the question of available learning guarantees regarding the model error on each distribution. \citet{shai2004} introduced the notion of $\mathcal{H}$-divergence to measure the distance between source and target marginal distributions in DA. \citet{shai2006,shai2010} have shown that a finite sample estimate of this divergence can be used to bound the target risk of the learned model.

The contributions of our work are threefold. First, we extend the DA guarantees to MDL (\sec~ \ref{sec:motivation}), showing that the risk of the learned model over all considered domains is upper bounded by the oracle risk and the sum of the $\mathcal{H}$-divergences between any two domains. Furthermore,  an upper bound on the classifier imbalance (the difference between the individual domain risk, and the average risk over all domains) is obtained, thus bounding the worst-domain risk. Second, we propose the approach {\em Multi-domain Learning Adversarial Neural Network} (\XX), which extends Domain Adversarial Neural Networks (\DANN s)~\citep{ganin} to semi-supervised DA and MDL. Relaxing the DA assumption, \XX\ handles the so-called class asymmetry issue (when each domain may contain varying numbers of labeled and unlabeled examples of a subset of all possible classes), through designing a new loss (\sec~\ref{sec:mulann}). Finally, \XX\ is empirically validated in both DA and MDL settings (\sec~\ref{results}), as it significantly outperforms the state of the art on three standard image benchmarks~\citep{office, mnist}, and a novel bioimage benchmark, \pheno, where the state of the art involves extensive domain-dependent pre-processing.

\paragraph{Notation.} Let $\mathcal{X}$ denote an input space and $\mathcal{Y} = \{ 1, \dots, L\}$ a set of classes. For $i=1,\ldots, n$, dataset $S_i$ is an iid sample drawn from distribution $\mathcal D_{i}$ on $\mathcal{X} \times \mathcal{Y}$. The marginal distribution of $\mathcal D_{i}$ on $\mathcal{X}$ is denoted by $\mathcal D_{i}^{\mathcal{X}}$. Let $\cal H$ be a hypothesis space; for each $h$ in $\cal H$ ($h: {\cal X} \mapsto \mathcal{Y}$) we define the risk under distribution ${\cal D}_{i}$ as $\epsilon_i(h) =
\mathbb P_{ \mathbf{x},y\sim \mathcal D_i}(h(\mathbf{x}) \neq y)$. $h_i^\star$ (respectively $h^\star$) denotes the oracle hypothesis according to distribution ${\cal D}_i$ (resp.  with minimal total risk over all domains): 
\begin{align}
\epsilon_i^\star =& \epsilon_i (h_i^\star) = \underset{h \in \mathcal H}{min}~ \epsilon_i(h)\\
\bar{\epsilon}(h^\star) =&\underset{h \in \mathcal H}{min}~\bar{\epsilon}(h) = \underset{h \in \mathcal H}{min}~\tfrac{1}{n} \underset{i}{\sum} \epsilon_i(h)
\end{align}

In the semi-supervised setting, the label associated with an instance might be missing. In the following, "domain" and "distribution" will be used interchangeably, and the "classes of a domain" denote the classes for which labeled or unlabeled examples are available in this domain. 

\vspace{-5pt}
\section{State of the art}
\label{soa}
\vspace{-5pt}
Machine learning classically relies on the iid setting: when training and test samples are independently drawn from the same joint distribution $P(X, Y)$ ~\citep{Vapnik98}. Two other settings emerged in the 1990s, "concept drift" and "covariate shift". They respectively occur when conditional data distributions $P(Y|X)$ and marginal data distributions $P(X)$ change, either continuously or 
abruptly, across training data or between train and test data~\citep{Shimodaira00}. Since then, transfer learning has come to designate methods to learn across drifting, shifting or distinct distributions, or even distinct tasks~\citep{PrattAAAI91,5288526}. Restricting ourselves to addressing a single task on a common input space, we distinguish two objectives:
minimizing the learning risk over {\em all} considered distributions (MDL), or over {\em a single} target distribution while exploiting samples from richer source(s) (DA). MDL is thus distinct from multiple source DA by their respective focus on the average risk over all distributions, versus target accuracy only. Samples from the different domains can be all, partially, or not labeled (supervised, semi-supervised and unsupervised settings). Finally, different domains can involve the same classes, or some domains can involve classes not included in other domains, referred to as {\em class asymmetry}.

In MDL, the different domains can be taken into account by maintaining shared and domain-specific parameters~\citep{Dredze2010}, or through a domain-specific use of shared parameters. The domain-dependent use of these parameters can be learned, e.g. using domain-guided dropout~\citep{Xiao2016}, or based on prior knowledge about domain semantic relationships~\citep{Yang2015}.

Early DA approaches leverage source examples to learn on the target domain in various ways, e.g. through reweighting source datapoints \citep{Mansour09,Huang:2006:CSS:2976456.2976532,Gong:2013:CDL:3042817.3042844}, 
or defining an extended representation to learn from both source and target~\citep{DaumeM06}. Other approaches proceed by aligning the source and target representations with PCA-based correlation alignment~\citep{coral}, or subspace alignment~\citep{Fernando15}. In the field of computer vision, a somewhat related way of mapping examples in one domain onto the other is image-to-image translation, possibly in combination with a generative adversarial network (see references in Appendix~\ref{sec:extended_soa}). 

Intuitively, the difficulty of DA crucially depends on the distance between source and target distribution
. Accordingly, a large set of DA methods proceed by reducing this distance in the original input space $\cal X$, e.g. via importance sampling~\citep{Scheffer} or
by modifying the source representation using optimal transport~\citep{CourtyFlamary,Courty18}
. Another option is to map source and target samples on a latent space where they will have minimal distance. Neural networks have been intensively exploited to build such latent spaces, either through generative adversarial mechanisms~\citep{DBLP:journals/corr/TzengHSD17,Ghifary16}, or through combining task objective with an approximation of the distance between source(s) and target. Examples of used distances include the Maximum Mean Discrepancy due to~\citet{Gretton07}~\citep{DBLP:journals/corr/TzengHZSD14, DBLP:journals/corr/BousmalisTSKE16}, some of its variants~\citep{jordan2015,Long_variante_MMD}, the $\mathcal L_2$ contrastive divergence \citep{motiian2017}, the Frobenius norm of the output feature correlation matrices \citep{deep_coral}, or the $\mathcal{H}$-divergence~\citep{shai2006,shai2010,ganin,mada,justanotherlongpaper} (more in \sec~\ref{theory}). 
Most DA methods assume that source(s) and target contain examples from the same classes;
in particular, in standard benchmarks such as \office~\citep{office}, all domains contain examples from the same classes. Notable exceptions are partial DA methods, where target classes are expected to be a subset of source classes e.g.~\citep{partialDA2,Cao_2018_partialDA}. DA and partial DA methods share two drawbacks when applied to semi-supervised MDL with non-identical domain class sets. First, neither generic nor partial DA methods try to mitigate the impact of unlabeled samples from a class without any labeled counterparts. Second, as they focus on target performance, (partial) DA methods do not discuss the impact of extra labeled source classes on source accuracy. However, as shown in \sec~\ref{sec:asym}, class asymmetry can heavily impact model performance if not accounted for.

Bioinformatics is increasingly appreciating the need for domain adaptation methods \citep{borgwardt2006,Schweikert:2008:EAD:2981780.2981959,Xu:2011fj,Vallania206466}. Indeed, experimentalists regularly face the issues of concept drift and covariate shift. Most biological experiments that last more than a few days are subject to technical variations between groups of samples, referred to as {\em batch effects}. Batch effects in image-based screening data are usually tackled with specific normalization methods~\citep{pmid19644458}. More recently, work by~\citet{Ando} applied CorAl~\citep{coral} for this purpose, aligning each batch with the entire experiment.
DA has been applied to image-based datasets for improving or accelerating image segmentation tasks~\citep{becker2015, opbroek2015,bermudez2016,kamnitsas2017}. However, to our knowledge, MDL has not yet been used in Bioimage Informatics, and this work is the first to leverage distinct microscopy screening datasets using MDL.

\vspace{-5pt}
\section{Multi-Domain Adversarial Learning}
\vspace{-5pt}
\label{theory}
The $\mathcal{H}$-divergence has been introduced to bound the DA risk~\citep{shai2006,shai2010,ganin}. This section extends the DA theoretical results to the MDL case (\sec~\ref{sec:motivation}), supporting the design of the \XX\ approach (\sec~\ref{sec:mulann}). The reader is referred to Appendix \ref{proof} for formal definitions and proofs. 
\vspace{-7pt}
\subsection{$\mathcal H$-divergence for MDL}
\label{sec:motivation}
\vspace{-7pt}
The distance between source and target partly governs the difficulty of DA.
The $\mathcal H$-divergence has been introduced to define such a distance which can be empirically estimated with proven guarantees~\citep{batu, shai2004}. This divergence measures how well one can discriminate between samples from two marginals. It inspired an adversarial approach to DA \citep{ganin}, through the finding of a feature space in which a binary classification loss between source and target projections is maximal, and thus their $\mathcal{H}$-divergence minimal. Furthermore, the target risk is upper-bounded by the empirical source risk, the empirical $\mathcal H$-divergence between source(s) and target marginals, and the oracle DA risk~\citep{shai2006,shai2010,NIPS2012_4684}. 

\vspace{-7pt}
\paragraph{Bounding the MDL loss using the $\mathcal H$-divergence.} A main difference between DA and MDL is that MDL aims to minimize the average risk over all domains %
while DA aims to minimize the target risk only. Considering for simplicity a binary classification MDL problem and taking inspiration from \citep{Mansour:2008:DAM:2981780.2981910,shai2010}, the MDL loss can be formulated as an optimal convex combination of domain risks. A straightforward extension of~\citet{shai2010} (Theorem~\ref{theorem1} in Appendix~\ref{sec:thm1}) establishes that the compound empirical risk is upper bounded by the sum of: i) the oracle risk on each domain; ii) a statistical learning term involving the VC dimension of $\mathcal{H}$; iii) the divergence among any two domains as measured by their $\mathcal{H}$-divergence and summed oracle risk. This result states that, assuming a representation in which domains are as indistinguishable as possible and on which every 1- and 2-domain classification task is well addressed, then there exists a model that performs well on all of them. In the 2-domain case, the bound is minimized when one minimizes the convex combination of losses in the same proportion as samples.

\vspace{-7pt}
\paragraph{Bounding the worst risk.}
The classifier imbalance w.r.t. the $i$-th domain is defined as $\vert \epsilon_i(h) - \bar{\epsilon}(h)\vert$. The extent to which marginal $\mathcal{D}_i$ can best be distinguished by a classifier from $\mathcal{H}$ (i.e., the $\mathcal{H}$-divergence), and the intrinsic difficulty $\epsilon^\star_i$ of the $i$-th classification task, yield an upper-bound on the classifier imbalance (proof in Appendix \ref{sec:proofProp2}):
\begin{prop}
\label{prop1}
    Given an input space $\mathcal X$, $n$ distributions $\mathcal D_i$ over $\mathcal X \times \{0,1\}$ and hypothesis class $\mathcal H$ on $\mathcal X$, for any $h \in \mathcal H$, let $\epsilon_i(h)$ (respectively $\bar{\epsilon}(h)$) denote
    the  classification risk of $h$ w.r.t. distribution ${\cal D}_i$ (resp. its average risk over all ${\cal D}_i$). The risk imbalance $\vert \epsilon_i(h) - \bar{\epsilon}(h)\vert$ is upper bounded as:
    \begin{equation}
    \vert \epsilon_i(h) - \bar{\epsilon}(h)\vert \leq \epsilon_i^\star + \dfrac{1}{n} \underset{j}{\sum} \epsilon_j^\star + \dfrac{1}{n} \underset{j}{\sum} \left( d_{\mathcal H} (\mathcal D_i^X, \mathcal D_j^X) + \Delta_{ij}\right)
    \end{equation}
    with $\Delta_{ij} = \max(E_{\mathcal D_j^X}\vert h_i^\star(\mathbf{x}) - h_j^\star(\mathbf{x})\vert, E_{\mathcal D_i^X}\vert h_i^\star(\mathbf{x}) - h_j^\star(\mathbf{x})\vert) $
\end{prop}
Accordingly, every care taken to minimize $\mathcal{H}$-divergences or $\Delta_{ij}$ (e.g. using the class-wise contrastive losses~\citep{motiian2017}) improves the above upper bound. An alternative bound of the classifier imbalance can be obtained by using the $\mathcal{H}\Delta\mathcal{H}$-divergence (proposition~\ref{prop2}, and corollaries~\ref{coro3},~\ref{coro4} for the 2-domain case in Appendix).

\vspace{-7pt}
\subsection{\XX: Multi-domain adversarial learning}
\label{sec:mulann}
\vspace{-7pt}

As pointed out by e.g.~\cite{mada}, when minimizing the $\mathcal{H}$-divergence between two domains, a negative transfer can occur in the case of class asymmetry, when domains involve distinct sets of classes. For instance, if a domain has unlabeled samples from a class which is not present in the other domains, both global~\citep{ganin} and class-wise~\citep{mada} domain alignments will likely deteriorate at least one of the domain risks by putting the unlabeled samples close to labeled ones from the same domain. A similar issue arises if a domain has no (labeled or unlabeled) samples in classes which are represented in other domains. In general, unlabeled samples are only subject to constraints from the domain discriminator, as opposed to labeled samples. Thus, in the case of class asymmetry, domain alignment will tend to shuffle unlabeled samples 
more than labeled ones.

\begin{figure}
\vspace{-35pt}
    \centerline{
    \includegraphics[width=0.65\linewidth]{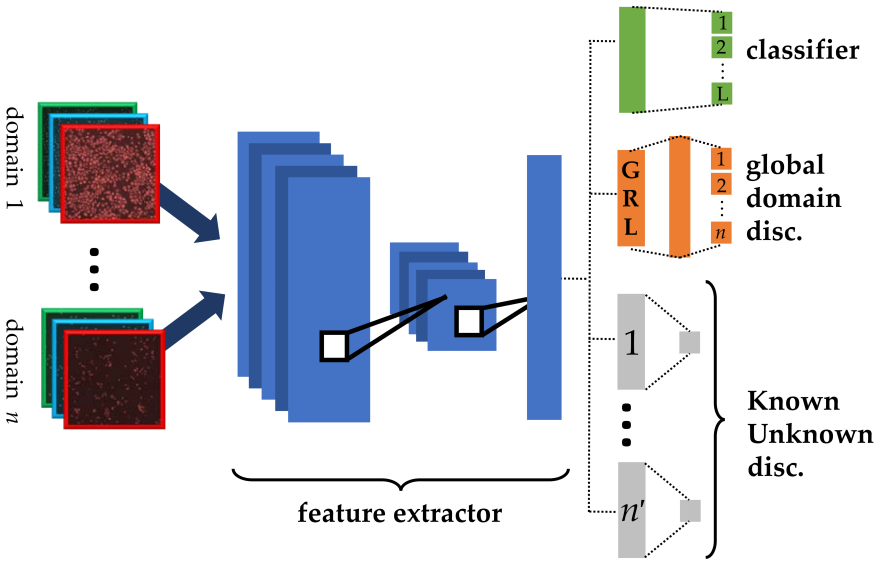}
    \includegraphics[width=0.35\linewidth]{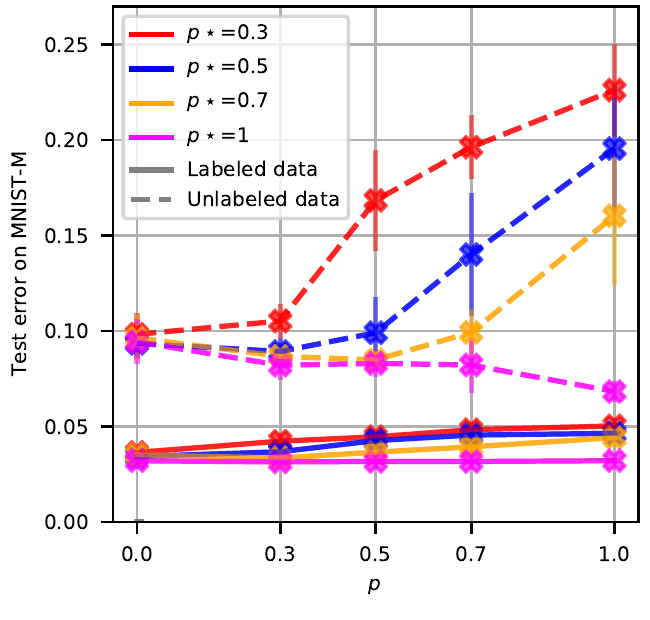}}
    \caption{Left: \XX\ architecture. GRL: gradient reversal layer from~\citet{ganin}. Right: impact of parameter $p$ in comparison with the groundtruth $p^\star$  on MNIST $\rightarrow$ MNIST-M. $p=0$ corresponds to DANN: no data flowed through the KUD module (see text for details).}
    \label{fig:architecture}
\end{figure}

This limitation is addressed in \XX\ by 
defining a new discrimination task
referred to as {\em Known Unknown Discrimination} (KUD). Let us assume that, in each domain, a fraction 
$p^\star$ of unlabeled samples comes from extra classes, i.e. classes with no labeled samples within the domain. KUD aims at discriminating, within each domain, labeled samples from unlabeled ones that most likely belong to such extra classes. 
More precisely, unlabeled samples of each domain are ranked according to the entropy of their classification according to the current classifier, restricted to their domain classes. Introducing the hyper-parameter $p$, the top $p$\% examples according to this classification entropy are deemed "most likely unknown", and thus discriminated from the labeled ones of the same domain.
The KUD module 
aims at repulsing the most likely unknown unlabeled samples from the labeled ones within each domain (Fig. \ref{fig:architecture}), thus resisting the contractive effects of global domain alignment.

Overall, \XX\ involves 3+$n'$ interacting modules, where $n'$ is the number of domains with unlabeled data. The first module is the feature extractor with parameters $\theta_f$, which maps the input space ${\cal X}$ to some latent feature space $\Omega$. 2+$n'$ modules are defined  on $\Omega$: the classifier module, the domain discriminator module, and the $n'$ KUD modules, with respective parameters $\theta_c$,  $\theta_d$ and $(\theta_{u,i})_i$. All modules are simultaneously 
learned
by minimizing loss $\mathcal{L}(\theta_f, \theta_c, \theta_d, \theta_{u})$:
\begin{equation}
\label{equation:mulann}
    \mathcal{L}(\theta_f, \theta_c, \theta_d, \theta_{u}) = \dfrac{1}{n} \sum_{i=1}^n \left( \mathcal{L}_c^{i} (\theta_f, \theta_c) - \lambda~ \mathcal{L}_d^{i} (\theta_f, \theta_d)\right) + \dfrac{\zeta}{n'} \sum_{j=1}^{n'} \mathcal{L}_u^{j}(\theta_f, \theta_{u,j}) \end{equation}
where $\zeta$ and $\lambda$ are hyper-parameters, $\mathcal{L}_c^{i} (\theta_f, \theta_c)$ is the empirical classification loss on labeled examples in $S_i$, $\mathcal{L}_d^{i} (\theta_f, \theta_d)$ is the domain discrimination loss (multi-class cross-entropy loss of classifying examples from $S_i$ in class $i$), and $\mathcal{L}_{u}^{i}(\theta_f, \theta_{u,i})$ is the KUD loss (binary cross-entropy loss of 
discriminating labelled samples from $S_i$ from the "most likely unknown" unlabelled samples from $S_i$). 

The loss minimization aims to find a saddle point $(\hat{\theta}_f, \hat{\theta}_y, \hat{\theta}_d, \hat{\theta}_u)$, achieving an equilibrium between the classification performance, the discrimination among domains (to be prevented) and the discrimination among labeled and some unlabeled samples within each domain (to be optimized). 
The sensitivity w.r.t. hyperparameter $p$ will be discussed in \sec~\ref{sec:asym}. 

\vspace{-5pt}
\section{Experimental validation}
\label{results}
\vspace{-5pt}
This section reports on the experimental validation of \XX\ in DA and MDL settings on three image datasets (\sec~\ref{sec:evaluation}), prior to analyzing \XX\ and investigating the impact of 
class asymmetry on model performances (\sec~\ref{sec:asym}).

\vspace{-7pt}
\subsection{Implementation}
\label{sec:implementation}
\vspace{-7pt}
\paragraph{Datasets} The DA setting considers three benchmarks: \digits, including the well-known MNIST and MNIST-M~\citep{mnist,ganin}; Synthetic road signs and German traffic sign benchmark~\citep{synsigns,gtsrb} and \office~\citep{office}
. The MDL setting considers the new  \pheno\ benchmark, which is made of fluorescence microscopy images of cells (detailed in Appendix~\ref{biodata}). Each image contains tens to hundreds of cells that have been exposed to a given chemical compound, in three domains: California (C), Texas (T) and England (E). 
There are 13 classes across the three domains (Appendix, Fig.~\ref{fig:image_bio}); a drug class is a group of compounds targeting a similar known biological process, e.g. DNA replication. Four domain shifts are considered: C$\leftrightarrow$T, T$\leftrightarrow$E, E$\leftrightarrow$C and C$\leftrightarrow$T$\leftrightarrow$E
.
\def\MADA{{\sc Mada}}
\vspace{-7pt}
\paragraph{Baselines and hyperparameters.}
In all experiments, \XX\ is compared to \DANN\ \citep{ganin} and its extension \MADA~\citep{mada} (that involves one domain discriminator module per class rather than a single global one). 
For \DANN, \MADA\ and \XX, the same pre-trained VGG-16 architecture~\citep{vgg16} from Caffe~\citep{caffe} is used  for \office\ and \pheno\footnote{Complementary experiments with AlexNet~\citep{alexnet} yield worse results, as already noted by~\citep{DBLP:journals/corr/KoniuszTP16}.}; the same small convolutional network as~\citet{ganin} is used for \digits\ (see Appendix~\ref{sec:architecture} for details). 
The models are trained in Torch~\citep{torch} using stochastic gradient descent with momentum ($\rho=0.9$). As in \citep{ganin}, no hyper-parameter grid-search is performed for \office\ results - double cross-validation is used for all other benchmarks. Hyper-parameter ranges can be found in Appendix~\ref{sec:hyperparam}.

\vspace{-7pt}
\paragraph{Semi-supervised setting.} For \office\ and \pheno, we follow the experimental settings from~\citet{office}. A fixed number of labeled images per class is used for one of the domains in all cases (20 for Amazon, 8 for DSLR and Webcam, 10 in \pheno). For the other domain, 10 labeled images per class are used for half of the classes (15 for \office, 4 for \pheno). For \digits\ and RoadSigns, all labeled source train data is used, whereas labeled target data is used for half of the classes only (5 for \digits, 22 for RoadSigns). In DA, the evaluation is performed on all target images from the unlabeled classes. In MDL, the evaluation is performed on all source and target classes (considering labeled and unlabeled samples).

\vspace{-7pt}
\paragraph{Evaluation goals.} A first goal is to assess \XX\ performance comparatively to the baselines. A second goal is to assess how the experimental setting impacts model performance. As domain discriminator and KUD modules can use both labeled and unlabeled images, a major question regards the impact of seeing unlabeled images during training. Two experiments are conducted to assess this impact: a) the same unlabeled images are used for training and evaluation (referred to as \textbf{fully transductive} setting, noted FT) ; b) some unlabeled images are used for training, and others for evaluation (referred to as \textbf{non-fully transductive} setting, noted NFT). (The case where no unlabeled images are used during training is discarded due to poor results).

\vspace{-7pt}
\subsection{Evaluation}
\label{sec:evaluation}
\vspace{-7pt}
\paragraph{DA on \digits, RoadSigns and \office.}
Table~\ref{table:office} compares \XX\ with \DANN\ and \MADA\ (\sec\ \ref{sec:implementation}). Other baselines include: Learning from source and target examples with no transfer loss; Published results from \citep{motiian2017} (legend CCSA), that uses a contrastive loss to penalizes large (resp. small) distances between same (resp. different) classes and different domains in the feature space; Published results from~\citep{7410820}, an extension of \DANN\ that adds a loss on target softmax values ("soft label loss"; legend Tseng15).
Overall, \XX\ yields the best results, significantly improving upon the former best results on the most difficult cases, i.e., D$\rightarrow$A, A$\rightarrow$D or W$\rightarrow$A. As could be expected, the fully transductive results match or significantly outperform the non-fully transductive ones. Notably, \MADA\ performs similarly to \DANN\ on \digits\ and RoadSigns, but worse on \office; a potential explanation is that \MADA\ is hindered as the number of classes, and thus domain discriminators, increases (respectively 10, 32 and 43 classes).

 \begin{table}
   \caption{Classification results on target test set in the semi-supervised DA setting (average and stdev on 5 seeds or folds). Bold: results less than 1 stdev from the best in each column. See text.}
   \label{table:office}
   \tabcolsep=0.1cm
   \small
   \begin{center}
   \begin{tabular}{llll|llllllll}
     \toprule
     \multicolumn{2}{l}{Source}& Mnist & SynSigns& DSLR & Amazon & Webcam & DSLR & Webcam & Amazon &\office \\
     \multicolumn{2}{l}{Target}& Mnist-M & GTSRB &Amazon & DSLR & DSLR & Webcam & Amazon & Webcam & average \\
      \midrule
    \multicolumn{2}{l}{Baseline} &35.6~(0.6)&85.1~(1.2)&35.5~(0.5)&58.5~(1.7)&90.9~(1.8)&90.6~(0.6)&34.4~(2.7)&55.8~(1.5)&61.0\\
     \midrule
     \multicolumn{2}{l}{Tzeng15}&-&-&43.1~(0.2)&68.0~(0.5)&97.5~(0.1)&90.0~(0.2)&40.5~(0.2)&59.3~(0.6)&66.4\\
     \multicolumn{2}{l}{CCSA} &-&-& 42.6~(0.6) & 70.5~(0.6) & 96.2~(0.3) & 90.0~(0.2) & 43.6~(1.0) & 63.3~(0.9) & 67.8\\
    \multirow{3}{*}{NFT} & \DANN &90.4~(1.1)&89.8~(1.1)&50.9~(2.4)&68.6~(4.9)&88.8~(3.2)&\textbf{91.9}~(0.7)&48.8~(3.8)&73.0~(2.6)&70.3\\
    &MADA &89.9~(0.8)&88.7~(1.0)&44.8~(3.3)&64.0~(3.9)&88.2~(4.2)&89.1~(3.4)&44.7~(4.8)&72.2~(3.1)&67.2\\
    &\XX &\textbf{91.5~(0.4)}&\textbf{92.1~(1.4)}&\textbf{57.6}~(3.9)&\textbf{75.8}~(3.7)&\textbf{93.3}~(2.5)&89.9~(1.6)&\textbf{54.9}~(3.9)&\textbf{76.8}~(3.1)&\textbf{74.7}\\
    \midrule
     \multirow{3}{*}{FT} &\DANN &90.6~(1.2)&86.7~(0.8)&52.2~(2.2)&77.4~(2.2)&\textbf{94.6}~(1.2)&\textbf{90.7}~(1.7)&53.0~(1.9)&74.3~(2.7)&73.7\\
    &MADA &91.0~(1.1)&84.8~(1.6)&51.6~(2.5)&78.8~(3.6)&91.7~(1.7)&88.8~(2.3)&53.8~(2.6)&73.5~(2.2)&73.0\\
    &\XX &\textbf{92.7~(0.6)}&\textbf{89.1~(1.5)}&\textbf{63.9}~(2.4)&\textbf{81.7}~(1.7)&\textbf{95.4}~(2.4)&\textbf{89.3}~(2.8)&\textbf{64.2}~(2.5)&\textbf{80.8}~(2.7)&\textbf{79.2}\\

     \bottomrule
   \end{tabular}
  \end{center}
 \end{table}

\vspace{-7pt}
\paragraph{MDL on \pheno.} A state of the art method for fluorescence microscopy images relies on tailored approaches for quantifying changes to cell morphology ~\citep{pmid26655497}. Objects (cells) are segmented in each image, and circa 650 shape, intensity and texture features are extracted for each object in each image. The {\em profile} of each image is defined as the vector of its Kolmogorov-Smirnov statistics, computed for each feature by comparing its distribution to that of the same feature from pooled negative controls of the same plate\footnote{A plate contains between 96 and 384 experiments, realized the same day in exactly the same conditions.}. Classification in profile space is realized using linear discriminant analysis, followed by k-nearest neighbor (LDA+k-NN) ("Baseline P" in Table~\ref{table:bio}). As a state of the art shallow approach to MDL to be applied in profile space, CORAL~\citep{coral} was chosen ("P + CORAL" in Table~\ref{table:bio}). A third baseline corresponds to fine-tuning VGG-16 without any transfer loss ("Baseline NN").

Table~\ref{table:bio} compares \DANN, \MADA\ and \XX\ to the baselines, where columns 4-7 (resp. 8-9) consider raw images (resp. the profile representations).\footnote{We could not obtain results with CCSA~\citep{motiian2017} on unlabeled classes.
} The fact that a profile-based baseline generally outperforms an image-based baseline was expected, as profiles are designed to reduce the impact of experimental settings (column 4 vs. 8). The fact that standard deviations tend to be larger here than for \office, RoadSigns or \digits\ is explained by a higher intra-class heterogeneity; some classes comprise images from different compounds with similar but not identical biological activity. Most interestingly, \XX\ and P+CORAL both improve classification accuracy on unlabeled classes at the cost of a slighty worse classification accuracy for the labeled classes (in all cases but one). This is explained as reducing the divergence between domain marginals on the latent feature space prevents the classifier from exploiting dataset-dependent biases. Overall, \XX\ and P+CORAL attain comparable results on two-domain cases, with \XX\ performing significantly better in the three-domain case. Finally, \XX\ matches or significantly outperforms \DANN\ and \MADA.
 \begin{table}
   \caption{\pheno\ test classification accuracy results on all domains (average and stdev on 5 folds), in the fully transductive setting (see table~\ref{table:bio_non_transductive} in Appendix for non-transductive ones, and sections~\ref{sec:bio_images},~\ref{sec:bio_classes} for details about image and class selection).}
   \label{table:bio}
   \centering
   \tabcolsep=0.1cm
   \small
   \begin{tabular}{lllllll|ll}
     \toprule
      Shift & Image set & \# classes & Baseline NN & \DANN & MADA & \XX & {Baseline P} & P+Coral
      \\
     \midrule
    \multirow{3}{*}{E-C}&E&7&63.7~(7.0)&62.9~(7.6)&59.5~(9.5)&64.4~(8.0)&74.1~(3.9)&58.4~(6.1)\\ 
    &C lab.&4&97.0~(1.6)&86.4~(10.3)&86.1~(6.5)&82.4~(10.2)&95.4~(3.2)&86.6~(6.0)\\ 
    &C unlab.&3&0.6~(1.2)&54.4~(18.3)&33.6~(17.5)&58.4~(19.7)&25.5~(5.7)&42.2~(9.5)\\ 
    \midrule 
    \multirow{3}{*}{C-T}&C&10&90.4~(1.8)&90.0~(1.3)&87.2~(2.4)&88.0~(3.6)&96.1~(1.0)&93.8~(0.9)\\ 
    &T lab.&7&93.8~(2.0)&93.6~(1.8)&89.2~(2.4)&90.0~(1.9)&95.2~(3.1)&93.4~(3.0)\\ 
    &T unlab.&3&36.4~(10.7)&68.3~(6.4)&63.7~(10.4)&91.6~(5.7)&68.1~(2.1)&86.0~(7.8)\\ 
    \midrule 
    \multirow{3}{*}{T-E}&T&7&88.9~(6.6)&90.8~(3.9)&87.7~(2.1)&85.7~(6.6)&89.3~(8.7)&90.3~(3.1)\\ 
    &E lab.&4&60.0~(5.3)&59.4~(6.8)&56.5~(12.3)&54.5~(6.5)&59.4~(8.1)&50.3~(6.4)\\ 
    &E unlab.&3&19.0~(14.4)&72.7~(10.1)&56.2~(16.6)&71.7~(21.9)&32.9~(12.3)&48.1~(10.0)\\ 
    \midrule 
   \multirow{3}{*}{C-T-E}&C&7&89.8~(3.5)&87.8~(4.6)&92.8~(1.5)&88.8~(5.2)&96.3~(1.1)&89.3~(5.0)\\ 
&T&7&92.6~(2.6)&90.2~(1.2)&94.2~(2.3)&92.5~(3.0)&96.8~(2.5)&89.9~(3.1)\\ 
&E lab.&4&62.3~(5.5)&56.7~(4.2)&53.6~(8.5)&48.1~(5.3)&57.3~(6.1)&44.4~(7.2)\\ 
&E unlab.&3&19.9~(13.5)&49.4~(6.5)&46.5~(6.9)&79.4~(5.3)&45.5~(13.6)&62.8~(7.2)\\ 
\bottomrule
   \end{tabular}
 \end{table}

\vspace{-7pt}
\subsection{Analyses} 
\label{sec:asym}
\vspace{-7pt}
Two complementary studies are conducted to investigate the impact of hyperparameter $p$ and that of class asymmetry. The tSNE~\citep{tsne} visualizations of the feature space for \DANN, \MADA\ and \XX\ are displayed in Appendix, Fig.~\ref{fig:tsne_large}.

\vspace{-7pt}
\paragraph{Sensitivity w.r.t. the fraction $p$ of "known unknowns".} \XX\ was designed to counter the negative transfer that is potentially caused by class asymmetry. This is achieved through the repulsion of labeled examples in each domain from the fraction $p$ of unlabeled examples deemed to belong to extra classes (not represented in the domain). The sensitivity of \XX\ performance to the value of $p$ and its difference to the ground truth $p^\star$ is investigated on MNIST$\leftrightarrow$MNIST-M. A first remark is that discrepancies between $p$ and $p^\star$ has no influence on the accuracy on a domain without unlabeled datapoints (Fig.~\ref{fig:p_eval_source} in Appendix). Fig.~\ref{fig:architecture}, right, displays the error depending on $p$ for various values of $p^\star$. As could have been expected, it is better to underestimate than to overestimate $p^\star$; it is even better to slightly underestimate it than to get it right, as the entropy ranking of unlabeled examples can be perturbed by classifier errors. 

\vspace{-7pt}
\paragraph{Impact of class/domain asymmetry.}
Section~\ref{sec:evaluation} reports on the classification accuracy when all classes are represented in all domains of a given shift. In the general case however, the classes represented by the unlabeled examples are unknown, hence there might exist "orphan" classes, with labeled or unlabeled samples, unique to a single domain. The impact of such orphan classes, referred to as class asymmetry, is investigated in the 2-domain case. Four types of samples are considered (Table 3): A class might have labeled examples in both domains ($\alpha$), labeled in one domain and unlabeled in the other domain ($\beta$), labeled in one domain and absent in the other one (orphan $\gamma$), and finally unlabeled in one domain and absent in the other one (orphan $\delta$). The impact of the class asymmetry is displayed on Fig. 3, reporting the average classification accuracy of $\alpha, \beta$ classes on domain 1 on the x-axis, and classification accuracy of unlabeled $\beta$ classes on domain 2 on the y-axis, for \XX, \DANN\ and \MADA\ on \office\ (on \pheno\ in Fig.~\ref{fig:asymmetry_cell}, Appendix).

A clear trend is that adding labeled orphans $\gamma$ (case "2", Fig. 3) entails a loss of accuracy for all algorithms compared to the no-orphan reference (case "1").
This is explained as follows: on the one hand, the $\gamma$ samples are subject to the classifier pressure as all labeled samples; on the other hand, they must be shuffled with samples from domain 2 due to the domain discriminator(s) pressure. Thus, the easiest solution is to shuffle the unlabeled $\beta$ samples around, and the loss of accuracy on these $\beta$ samples is very significant (the "2" is lower on the $y$-axis compared to "1" for all algorithms). The perturbation is less severe for the labeled $(\alpha, \beta)$ samples in domain 1, which are preserved by the classifier pressure ($x$-axis). 

The results in case "3" are consistent with the above explanation: since the unlabeled $\delta$ samples are only seen by the discriminator(s), their addition has little impact on either the labeled or unlabeled data classification accuracy 
(Figs. 3 and~\ref{fig:asymmetry_cell}). Finally, there is no clear trend in the impact of both labeled and unlabeled orphans (case "4"): labeled $(\alpha, \beta)$ (resp. unlabeled $\beta$) are only affected for \MADA\ on \pheno\ (resp. \XX\ on \office). 
Overall, these results show that class asymmetry matters for practical applications of transfer learning, and can adversely affect all three adversarial methods (Figs. 3 and~\ref{fig:asymmetry_cell}), with asymmetry in labeled class content ("2") being the most detrimental to model performance. 

\begin{figure}
\begin{tabular}{ll}
\parbox[b]{4.5cm}{
    {\footnotesize{\setlength{\tabcolsep}{1pt}
       \begin{tabular}{l|l||c|l}
         \toprule
          \multirow{2}{*}{Case} & Dom. 1 & \multicolumn{2}{c}{Dom. 2}\\
               & Lab. & Lab. & Unlab. \\
         \midrule
         1 & ~$\alpha, \beta$ & $\alpha$ & ~$\beta$ \\
         2 & ~$\alpha, \beta, \gamma$ & $\alpha$ & ~$\beta$ \\
         3 & ~$\alpha, \beta$ & $\alpha$ & ~$\beta, \delta$ \\
         4 & ~$\alpha, \beta, \gamma$ & $\alpha$ & ~$\beta, \delta$ \\
    \bottomrule
       \end{tabular}
       ~\\~\\
    }} 
    } & 
    \includegraphics[width=0.4\linewidth]{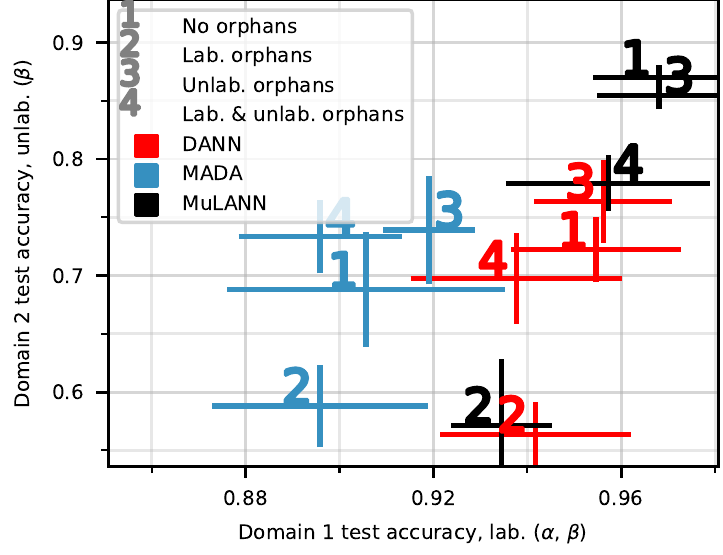}
 \\
 \parbox[b]{4.5cm}{Table 3: Class content per case in the asymmetry experiments}
 &  \parbox[b]{8.5cm}{Figure 3:
Impact of asymmetry in class content between domains on \office\ (W$\rightarrow$A) for \DANN, MADA and \XX. See text for details. Better seen in color.} 
    \label{fig:asym}
\end{tabular}
\end{figure}

\vspace{-5pt}
\section{Discussion and further work}
\vspace{-5pt}
\label{discussion}
This paper extends the use of domain adversarial learning to multi-domain learning, establishing how the $\mathcal{H}$-divergence can be used to bound both the risk across all domains and the worst-domain risk (imbalance on a specific domain). The stress is put on the notion of class asymmetry, that is, when some domains contain labeled or unlabeled examples of classes not present in other domains. Showing the significant impact of class asymmetry on the state of the art, this paper also introduces \XX, where a new loss is meant to resist the contractive effects of the adversarial domain discriminator and to repulse (a fraction of) unlabeled examples from labeled ones in each domain. The merits of the approach are satisfactorily demonstrated by comparison to \DANN\ and \MADA\ on \digits, RoadSigns and \office, and results obtained on the real-world \pheno\ problem establish a new baseline for the microscopy image community. 

A perspective for further study is to bridge the gap between the proposed loss and importance sampling techniques, iteratively exploiting the latent representation to identify orphan samples 
and adapt the loss while learning. Further work will also focus on how to identify and preserve relevant domain-specific behaviours while learning in a domain adversarial setting (e.g., if different cell types have distinct responses to the same class of perturbations). 

\subsubsection*{Acknowledgments}
This work was supported by NIH RO1 CA184984 (LFW), R01GM112690 (SJA) and the Institute of Computational Health Sciences at UCSF (SJA and LFW). We thank the Shoichet lab (UCSF) for access to their GPUs and Theresa Gebert for suggestions and feedback.
\bibliographystyle{iclr2019_conference}
{\small{\bibliography{bibliography}}}

\begin{thebibliography}{78}
\providecommand{\natexlab}[1]{#1}
\providecommand{\url}[1]{\texttt{#1}}
\expandafter\ifx\csname urlstyle\endcsname\relax
  \providecommand{\doi}[1]{doi: #1}\else
  \providecommand{\doi}{doi: \begingroup \urlstyle{rm}\Url}\fi

\bibitem[Ando et~al.(2017)Ando, McLean, and Berndl]{Ando}
D.~Michael Ando, Cory McLean, and Marc Berndl.
\newblock Improving phenotypic measurements in high-content imaging screens.
\newblock \emph{bioRxiv}, 2017.
\newblock \doi{10.1101/161422}.
\newblock URL \url{https://www.biorxiv.org/content/early/2017/07/10/161422}.

\bibitem[Anoosheh et~al.(2017)Anoosheh, Agustsson, Timofte, and
  Gool]{Anoosheh17}
Asha Anoosheh, Eirikur Agustsson, Radu Timofte, and Luc~Van Gool.
\newblock Combogan: Unrestrained scalability for image domain translation.
\newblock \emph{CoRR}, abs/1712.06909, 2017.

\bibitem[Batu et~al.(2000)Batu, Fortnow, Rubinfeld, Smith, and White]{batu}
Tugkan Batu, Lance Fortnow, Ronitt Rubinfeld, Warren~D. Smith, and Patrick
  White.
\newblock Testing that distributions are close.
\newblock In \emph{41st Annual Symposium on Foundations of Computer Science,
  {FOCS} 2000, 12-14 November 2000, Redondo Beach, California, {USA}}, pp.\
  259--269, 2000.
\newblock \doi{10.1109/SFCS.2000.892113}.
\newblock URL \url{https://doi.org/10.1109/SFCS.2000.892113}.

\bibitem[Becker et~al.(2015)Becker, Christoudias, and Fua]{becker2015}
C.~Becker, C.~M. Christoudias, and P.~Fua.
\newblock {{D}omain adaptation for microscopy imaging}.
\newblock \emph{IEEE Trans Med Imaging}, 34\penalty0 (5):\penalty0 1125--1139,
  May 2015.

\bibitem[Ben-David et~al.(2006)Ben-David, Blitzer, Crammer, and
  Pereira]{shai2006}
Shai Ben-David, John Blitzer, Koby Crammer, and Fernando Pereira.
\newblock Analysis of representations for domain adaptation.
\newblock In \emph{Proceedings of the 19th International Conference on Neural
  Information Processing Systems}, NIPS'06, pp.\  137--144, Cambridge, MA, USA,
  2006. MIT Press.
\newblock URL \url{http://dl.acm.org/citation.cfm?id=2976456.2976474}.

\bibitem[Ben-David et~al.(2010)Ben-David, Blitzer, Crammer, Kulesza, Pereira,
  and Vaughan]{shai2010}
Shai Ben-David, John Blitzer, Koby Crammer, Alex Kulesza, Fernando Pereira, and
  Jennifer~Wortman Vaughan.
\newblock A theory of learning from different domains.
\newblock \emph{Machine Learning}, 79\penalty0 (1):\penalty0 151--175, May
  2010.

\bibitem[Berm{\'u}dez-Chac{\'o}n et~al.(2016)Berm{\'u}dez-Chac{\'o}n, Becker,
  Salzmann, and Fua]{bermudez2016}
R{\'o}ger Berm{\'u}dez-Chac{\'o}n, Carlos Becker, Mathieu Salzmann, and Pascal
  Fua.
\newblock Scalable unsupervised domain adaptation for electron microscopy.
\newblock In Sebastien Ourselin, Leo Joskowicz, Mert~R. Sabuncu, Gozde Unal,
  and William Wells (eds.), \emph{Medical Image Computing and Computer-Assisted
  Intervention -- MICCAI 2016}, pp.\  326--334, Cham, 2016. Springer
  International Publishing.
\newblock ISBN 978-3-319-46723-8.

\bibitem[Bickel et~al.(2007)Bickel, Br\"{u}ckner, and Scheffer]{Scheffer}
Steffen Bickel, Michael Br\"{u}ckner, and Tobias Scheffer.
\newblock Discriminative learning for differing training and test
  distributions.
\newblock In \emph{Proceedings of the 24th International Conference on Machine
  Learning}, ICML '07, pp.\  81--88, New York, NY, USA, 2007. ACM.
\newblock ISBN 978-1-59593-793-3.
\newblock \doi{10.1145/1273496.1273507}.
\newblock URL \url{http://doi.acm.org/10.1145/1273496.1273507}.

\bibitem[Birmingham et~al.(2009)Birmingham, Selfors, Forster, Wrobel, Kennedy,
  Shanks, Santoyo-Lopez, Dunican, Long, Kelleher, Smith, Beijersbergen, Ghazal,
  and Shamu]{pmid19644458}
A.~Birmingham, L.~M. Selfors, T.~Forster, D.~Wrobel, C.~J. Kennedy, E.~Shanks,
  J.~Santoyo-Lopez, D.~J. Dunican, A.~Long, D.~Kelleher, Q.~Smith, R.~L.
  Beijersbergen, P.~Ghazal, and C.~E. Shamu.
\newblock {{S}tatistical methods for analysis of high-throughput {R}{N}{A}
  interference screens}.
\newblock \emph{Nat. Methods}, 6\penalty0 (8):\penalty0 569--575, Aug 2009.

\bibitem[Borgwardt et~al.(2006)Borgwardt, Gretton, Rasch, Kriegel, Scholkopf,
  and Smola]{borgwardt2006}
K.~M. Borgwardt, A.~Gretton, M.~J. Rasch, H.~P. Kriegel, B.~Scholkopf, and
  A.~J. Smola.
\newblock {{I}ntegrating structured biological data by {K}ernel {M}aximum
  {M}ean {D}iscrepancy}.
\newblock \emph{Bioinformatics}, 22\penalty0 (14):\penalty0 49--57, Jul 2006.

\bibitem[Bousmalis et~al.(2016)Bousmalis, Trigeorgis, Silberman, Krishnan, and
  Erhan]{DBLP:journals/corr/BousmalisTSKE16}
Konstantinos Bousmalis, George Trigeorgis, Nathan Silberman, Dilip Krishnan,
  and Dumitru Erhan.
\newblock Domain separation networks.
\newblock \emph{CoRR}, abs/1608.06019, 2016.
\newblock URL \url{http://arxiv.org/abs/1608.06019}.

\bibitem[Bousquet \& Elisseeff(2002)Bousquet and
  Elisseeff]{DBLP:journals/jmlr/BousquetE02}
Olivier Bousquet and Andr{\'{e}} Elisseeff.
\newblock Stability and generalization.
\newblock \emph{Journal of Machine Learning Research}, 2:\penalty0 499--526,
  2002.

\bibitem[Caie et~al.(2010)Caie, Walls, Ingleston-Orme, Daya, Houslay, Eagle,
  Roberts, and Carragher]{pmid20530715}
P.~D. Caie, R.~E. Walls, A.~Ingleston-Orme, S.~Daya, T.~Houslay, R.~Eagle,
  M.~E. Roberts, and N.~O. Carragher.
\newblock {{H}igh-content phenotypic profiling of drug response signatures
  across distinct cancer cells}.
\newblock \emph{Mol. Cancer Ther.}, 9\penalty0 (6):\penalty0 1913--1926, Jun
  2010.

\bibitem[Cao et~al.(2018)Cao, Long, Wang, and Jordan]{Cao_2018_partialDA}
Zhangjie Cao, Mingsheng Long, Jianmin Wang, and Michael~I. Jordan.
\newblock Partial transfer learning with selective adversarial networks.
\newblock In \emph{The IEEE Conference on Computer Vision and Pattern
  Recognition (CVPR)}, June 2018.

\bibitem[Chigorin et~al.(2012)Chigorin, Krivovyaz, Velizhev, and
  Konushin]{synsigns}
Alexander Chigorin, Gleb Krivovyaz, Alexander Velizhev, and Anton Konushin.
\newblock A method for traffic sign detection in an image with learning from
  synthetic data.
\newblock In \emph{14th International Conference Digital Signal Processing and
  its Applications}, volume~2, pp.\  316--319, 2012.
\newblock URL
  \url{http://graphics.cs.msu.ru/files/papers/dspa2012_chigorin_ts_recognition.pdf}.

\bibitem[Choi et~al.(2018)Choi, Choi, Kim, Ha, Kim, and Choo]{Choi17}
Yunjey Choi, Minje Choi, Munyoung Kim, Jung-Woo Ha, Sunghun Kim, and Jaegul
  Choo.
\newblock Stargan: Unified generative adversarial networks for multi-domain
  image-to-image translation.
\newblock In \emph{The IEEE Conference on Computer Vision and Pattern
  Recognition (CVPR)}, June 2018.

\bibitem[Collobert et~al.(2011)Collobert, Kavukcuoglu, and Farabet]{torch}
R.~Collobert, K.~Kavukcuoglu, and C.~Farabet.
\newblock Torch7: A matlab-like environment for machine learning.
\newblock In \emph{BigLearn, NIPS Workshop}, 2011.

\bibitem[Courty et~al.(2017)Courty, Flamary, Tuia, and
  Rakotomamonjy]{CourtyFlamary}
N.~Courty, R.~Flamary, D.~Tuia, and A.~Rakotomamonjy.
\newblock Optimal transport for domain adaptation.
\newblock \emph{IEEE Transactions on Pattern Analysis and Machine
  Intelligence}, 39\penalty0 (9):\penalty0 1853--1865, Sept 2017.
\newblock ISSN 0162-8828.
\newblock \doi{10.1109/TPAMI.2016.2615921}.

\bibitem[Damodaran et~al.(2018)Damodaran, Kellenberger, Flamary, Tuia, and
  Courty]{Courty18}
Bharath~Bhushan Damodaran, Benjamin Kellenberger, R{\'{e}}mi Flamary, Devis
  Tuia, and Nicolas Courty.
\newblock Deepjdot: Deep joint distribution optimal transport for unsupervised
  domain adaptation.
\newblock \emph{CoRR}, abs/1803.10081, 2018.
\newblock URL \url{http://arxiv.org/abs/1803.10081}.

\bibitem[Daum{\'{e}}~III \& Marcu(2006)Daum{\'{e}}~III and Marcu]{DaumeM06}
Hal Daum{\'{e}}~III and Daniel Marcu.
\newblock Domain adaptation for statistical classifiers.
\newblock \emph{J. Artif. Intell. Res.}, 26:\penalty0 101--126, 2006.

\bibitem[Dredze et~al.(2010)Dredze, Kulesza, and Crammer]{Dredze2010}
Mark Dredze, Alex Kulesza, and Koby Crammer.
\newblock Multi-domain learning by confidence-weighted parameter combination.
\newblock \emph{Machine Learning}, 79\penalty0 (1):\penalty0 123--149, May
  2010.
\newblock ISSN 1573-0565.
\newblock \doi{10.1007/s10994-009-5148-0}.
\newblock URL \url{https://doi.org/10.1007/s10994-009-5148-0}.

\bibitem[Fernando et~al.(2015)Fernando, Tommasi, and Tuytelaars]{Fernando15}
Basura Fernando, Tatiana Tommasi, and Tinne Tuytelaars.
\newblock Joint cross-domain classification and subspace learning for
  unsupervised adaptation.
\newblock \emph{Pattern Recognition Letters}, 65:\penalty0 60--66, 2015.
\newblock \doi{10.1016/j.patrec.2015.07.009}.
\newblock URL \url{https://doi.org/10.1016/j.patrec.2015.07.009}.

\bibitem[Ganin et~al.(2016)Ganin, Ustinova, Ajakan, Germain, Larochelle,
  Laviolette, Marchand, and Lempitsky]{ganin}
Yaroslav Ganin, Evgeniya Ustinova, Hana Ajakan, Pascal Germain, Hugo
  Larochelle, Fran\c{c}ois Laviolette, Mario Marchand, and Victor Lempitsky.
\newblock Domain-adversarial training of neural networks.
\newblock \emph{Journal of Machine Learning Research}, 17\penalty0
  (59):\penalty0 1--35, 2016.
\newblock URL \url{http://jmlr.org/papers/v17/15-239.html}.

\bibitem[Ghifary et~al.(2016)Ghifary, Kleijn, Zhang, Balduzzi, and
  Li]{Ghifary16}
Muhammad Ghifary, W.~Bastiaan Kleijn, Mengjie Zhang, David Balduzzi, and Wen
  Li.
\newblock Deep reconstruction-classification networks for unsupervised domain
  adaptation.
\newblock In \emph{Computer Vision - {ECCV}}, pp.\  597--613, 2016.

\bibitem[Gong et~al.(2013)Gong, Grauman, and
  Sha]{Gong:2013:CDL:3042817.3042844}
Boqing Gong, Kristen Grauman, and Fei Sha.
\newblock Connecting the dots with landmarks: Discriminatively learning
  domain-invariant features for unsupervised domain adaptation.
\newblock In \emph{Proceedings of the 30th International Conference on
  International Conference on Machine Learning - Volume 28}, ICML'13, pp.\
  I--222--I--230. JMLR.org, 2013.
\newblock URL \url{http://dl.acm.org/citation.cfm?id=3042817.3042844}.

\bibitem[Gretton et~al.(2007)Gretton, Borgwardt, Rasch, Sch{\"{o}}lkopf, and
  Smola]{Gretton07}
Arthur Gretton, Karsten~M. Borgwardt, Malte~J. Rasch, Bernhard Sch{\"{o}}lkopf,
  and Alexander~J. Smola.
\newblock A kernel method for the two-sample-problem.
\newblock In \emph{{NIPS}}, pp.\  513--520. {MIT} Press, 2007.

\bibitem[Huang et~al.(2006)Huang, Smola, Gretton, Borgwardt, and
  Scholkopf]{Huang:2006:CSS:2976456.2976532}
Jiayuan Huang, Alexander~J. Smola, Arthur Gretton, Karsten~M. Borgwardt, and
  Bernhard Scholkopf.
\newblock Correcting sample selection bias by unlabeled data.
\newblock In \emph{Proceedings of the 19th International Conference on Neural
  Information Processing Systems}, NIPS'06, pp.\  601--608, Cambridge, MA, USA,
  2006. MIT Press.
\newblock URL \url{http://dl.acm.org/citation.cfm?id=2976456.2976532}.

\bibitem[Isola et~al.(2017)Isola, Zhu, Zhou, and Efros]{Isola16}
Phillip Isola, Jun{-}Yan Zhu, Tinghui Zhou, and Alexei~A. Efros.
\newblock Image-to-image translation with conditional adversarial networks.
\newblock In \emph{{CVPR}}, pp.\  5967--5976. {IEEE} Computer Society, 2017.

\bibitem[Jia et~al.(2014)Jia, Shelhamer, Donahue, Karayev, Long, Girshick,
  Guadarrama, and Darrell]{caffe}
Yangqing Jia, Evan Shelhamer, Jeff Donahue, Sergey Karayev, Jonathan Long, Ross
  Girshick, Sergio Guadarrama, and Trevor Darrell.
\newblock Caffe: Convolutional architecture for fast feature embedding.
\newblock \emph{arXiv preprint arXiv:1408.5093}, 2014.

\bibitem[Jones et~al.(2001--)Jones, Oliphant, Peterson, et~al.]{scipy}
Eric Jones, Travis Oliphant, Pearu Peterson, et~al.
\newblock {SciPy}: Open source scientific tools for {Python}, 2001--.
\newblock URL \url{http://www.scipy.org/}.
\newblock [Online; accessed <today>].

\bibitem[Kamnitsas et~al.(2017)Kamnitsas, Baumgartner, Ledig, Newcombe,
  Simpson, Kane, Menon, Nori, Criminisi, Rueckert, and Glocker]{kamnitsas2017}
Konstantinos Kamnitsas, Christian Baumgartner, Christian Ledig, Virginia
  Newcombe, Joanna Simpson, Andrew Kane, David Menon, Aditya Nori, Antonio
  Criminisi, Daniel Rueckert, and Ben Glocker.
\newblock Unsupervised domain adaptation in brain lesion segmentation with
  adversarial networks.
\newblock In Marc Niethammer, Martin Styner, Stephen Aylward, Hongtu Zhu, Ipek
  Oguz, Pew-Thian Yap, and Dinggang Shen (eds.), \emph{Information Processing
  in Medical Imaging}, pp.\  597--609, Cham, 2017. Springer International
  Publishing.
\newblock ISBN 978-3-319-59050-9.

\bibitem[Kang et~al.(2016)Kang, Hsu, Wu, Liu, Coster, Posner, Altschuler, and
  Wu]{pmid26655497}
J.~Kang, C.~H. Hsu, Q.~Wu, S.~Liu, A.~D. Coster, B.~A. Posner, S.~J.
  Altschuler, and L.~F. Wu.
\newblock {{I}mproving drug discovery with high-content phenotypic screens by
  systematic selection of reporter cell lines}.
\newblock \emph{Nat. Biotechnol.}, 34\penalty0 (1):\penalty0 70--77, Jan 2016.

\bibitem[Kifer et~al.(2004)Kifer, Ben-David, and Gehrke]{shai2004}
Daniel Kifer, Shai Ben-David, and Johannes Gehrke.
\newblock Detecting change in data streams.
\newblock In \emph{Proceedings of the Thirtieth International Conference on
  Very Large Data Bases - Volume 30}, VLDB '04, pp.\  180--191. VLDB Endowment,
  2004.
\newblock ISBN 0-12-088469-0.
\newblock URL \url{http://dl.acm.org/citation.cfm?id=1316689.1316707}.

\bibitem[Koniusz et~al.(2016)Koniusz, Tas, and
  Porikli]{DBLP:journals/corr/KoniuszTP16}
Piotr Koniusz, Yusuf Tas, and Fatih Porikli.
\newblock Domain adaptation by mixture of alignments of second- or higher-order
  scatter tensors.
\newblock \emph{CoRR}, abs/1611.08195, 2016.
\newblock URL \url{http://arxiv.org/abs/1611.08195}.

\bibitem[Krizhevsky et~al.(2012)Krizhevsky, Sutskever, and Hinton]{alexnet}
Alex Krizhevsky, Ilya Sutskever, and Geoffrey~E Hinton.
\newblock Imagenet classification with deep convolutional neural networks.
\newblock In F.~Pereira, C.~J.~C. Burges, L.~Bottou, and K.~Q. Weinberger
  (eds.), \emph{Advances in Neural Information Processing Systems 25}, pp.\
  1097--1105. Curran Associates, Inc., 2012.
\newblock URL
  \url{http://papers.nips.cc/paper/4824-imagenet-classification-with-deep-convolutional-neural-networks.pdf}.

\bibitem[Le~Cun et~al.(1998)Le~Cun, Bottou, Bengio, and Haffner]{mnist}
Y.~Le~Cun, L.~Bottou, Y.~Bengio, and P.~Haffner.
\newblock Gradient-based learning applied to document recognition.
\newblock \emph{Proceedings of the IEEE}, 86\penalty0 (11):\penalty0
  2278--2324, Nov 1998.
\newblock ISSN 0018-9219.
\newblock \doi{10.1109/5.726791}.

\bibitem[Liu et~al.(2017)Liu, Breuel, and Kautz]{Liu17}
Ming-Yu Liu, Thomas Breuel, and Jan Kautz.
\newblock Unsupervised image-to-image translation networks.
\newblock In I.~Guyon, U.~V. Luxburg, S.~Bengio, H.~Wallach, R.~Fergus,
  S.~Vishwanathan, and R.~Garnett (eds.), \emph{Advances in Neural Information
  Processing Systems 30}, pp.\  700--708. Curran Associates, Inc., 2017.
\newblock URL
  \url{http://papers.nips.cc/paper/6672-unsupervised-image-to-image-translation-networks.pdf}.

\bibitem[Ljosa et~al.(2012)Ljosa, Sokolnicki, and Carpenter]{pmid22743765}
V.~Ljosa, K.~L. Sokolnicki, and A.~E. Carpenter.
\newblock {{A}nnotated high-throughput microscopy image sets for validation}.
\newblock \emph{Nat. Methods}, 9\penalty0 (7):\penalty0 637, Jun 2012.

\bibitem[Long et~al.(2015)Long, Cao, Wang, and Jordan]{jordan2015}
Mingsheng Long, Yue Cao, Jianmin Wang, and Michael~I. Jordan.
\newblock Learning transferable features with deep adaptation networks.
\newblock In \emph{Proceedings of the 32Nd International Conference on
  International Conference on Machine Learning - Volume 37}, ICML'15, pp.\
  97--105. JMLR.org, 2015.
\newblock URL \url{http://dl.acm.org/citation.cfm?id=3045118.3045130}.

\bibitem[Long et~al.(2016)Long, Wang, and Jordan]{Long_variante_MMD}
Mingsheng Long, Jianmin Wang, and Michael~I. Jordan.
\newblock Deep transfer learning with joint adaptation networks.
\newblock \emph{CoRR}, abs/1605.06636, 2016.
\newblock URL \url{http://arxiv.org/abs/1605.06636}.

\bibitem[Long et~al.(2017)Long, Cao, Wang, and Jordan]{justanotherlongpaper}
Mingsheng Long, Zhangjie Cao, Jianmin Wang, and Michael~I. Jordan.
\newblock Domain adaptation with randomized multilinear adversarial networks.
\newblock \emph{CoRR}, abs/1705.10667, 2017.
\newblock URL \url{http://arxiv.org/abs/1705.10667}.

\bibitem[Mansour(2009)]{Mansour09}
Yishay Mansour.
\newblock Learning and domain adaptation.
\newblock In \emph{Algorithmic Learning Theory, 20th International Conference,
  {ALT}}, pp.\  4--6, 2009.

\bibitem[Mansour et~al.(2008)Mansour, Mohri, and
  Rostamizadeh]{Mansour:2008:DAM:2981780.2981910}
Yishay Mansour, Mehryar Mohri, and Afshin Rostamizadeh.
\newblock Domain adaptation with multiple sources.
\newblock In \emph{Proceedings of the 21st International Conference on Neural
  Information Processing Systems}, NIPS'08, pp.\  1041--1048, USA, 2008. Curran
  Associates Inc.
\newblock ISBN 978-1-6056-0-949-2.
\newblock URL \url{http://dl.acm.org/citation.cfm?id=2981780.2981910}.

\bibitem[Motiian et~al.(2017)Motiian, Piccirilli, Adjeroh, and
  Doretto]{motiian2017}
Saeid Motiian, Marco Piccirilli, Donald~A. Adjeroh, and Gianfranco Doretto.
\newblock Unified deep supervised domain adaptation and generalization.
\newblock In \emph{The IEEE International Conference on Computer Vision
  (ICCV)}, Oct 2017.

\bibitem[Muandet et~al.(2013)Muandet, Balduzzi, and Sch\"{o}lkopf]{krikamol}
Krikamol Muandet, David Balduzzi, and Bernhard Sch\"{o}lkopf.
\newblock Domain generalization via invariant feature representation.
\newblock In \emph{Proceedings of the 30th International Conference on
  International Conference on Machine Learning - Volume 28}, ICML'13, pp.\
  I--10--I--18. JMLR.org, 2013.
\newblock URL \url{http://dl.acm.org/citation.cfm?id=3042817.3042820}.

\bibitem[Otsu(1979)]{otsu}
Nobuyuki Otsu.
\newblock {A} {T}hreshold {S}election {M}ethod from {G}ray-level {H}istograms.
\newblock \emph{IEEE Transactions on Systems, Man and Cybernetics}, 9\penalty0
  (1):\penalty0 62--66, 1979.
\newblock \doi{10.1109/TSMC.1979.4310076}.
\newblock URL \url{http://dx.doi.org/10.1109/TSMC.1979.4310076}.

\bibitem[Pan \& Yang(2010)Pan and Yang]{5288526}
S.~J. Pan and Q.~Yang.
\newblock A survey on transfer learning.
\newblock \emph{IEEE Transactions on Knowledge and Data Engineering},
  22\penalty0 (10):\penalty0 1345--1359, Oct 2010.
\newblock ISSN 1041-4347.

\bibitem[Pei et~al.(2018)Pei, Cao, Long, and Wang]{mada}
Zhongyi Pei, Zhangjie Cao, Mingsheng Long, and Jianmin Wang.
\newblock Multi-adversarial domain adaptation.
\newblock In \emph{Proceedings of the 32nd AAAI Conference on Artificial
  Intelligence}, 2018.
\newblock URL
  \url{https://www.aaai.org/ocs/index.php/AAAI/AAAI18/paper/view/17067}.

\bibitem[Pratt et~al.(1991)Pratt, Mostow, and Kamm]{PrattAAAI91}
L.~Y. Pratt, J.~Mostow, and C.~A. Kamm.
\newblock Direct transfer of learned information among neural networks.
\newblock In \emph{Proceedings of the Ninth National Conference on Artificial
  Intelligence (AAAI-91)}, AAAI'91, pp.\  584--589. Anaheim, CA, 1991.
\newblock URL \url{https://www.aaai.org/Papers/AAAI/1991/AAAI91-091.pdf}.

\bibitem[Preibisch et~al.(2009)Preibisch, Saalfeld, and Tomancak]{pmid19346324}
S.~Preibisch, S.~Saalfeld, and P.~Tomancak.
\newblock {{G}lobally optimal stitching of tiled 3{D} microscopic image
  acquisitions}.
\newblock \emph{Bioinformatics}, 25\penalty0 (11):\penalty0 1463--1465, Jun
  2009.

\bibitem[Russakovsky et~al.(2015)Russakovsky, Deng, Su, Krause, Satheesh, Ma,
  Huang, Karpathy, Khosla, Bernstein, Berg, and Fei-Fei]{ILSVRC15}
Olga Russakovsky, Jia Deng, Hao Su, Jonathan Krause, Sanjeev Satheesh, Sean Ma,
  Zhiheng Huang, Andrej Karpathy, Aditya Khosla, Michael Bernstein,
  Alexander~C. Berg, and Li~Fei-Fei.
\newblock {ImageNet Large Scale Visual Recognition Challenge}.
\newblock \emph{International Journal of Computer Vision (IJCV)}, 115\penalty0
  (3):\penalty0 211--252, 2015.
\newblock \doi{10.1007/s11263-015-0816-y}.

\bibitem[Saenko et~al.(2010)Saenko, Kulis, Fritz, and Darrell]{office}
Kate Saenko, Brian Kulis, Mario Fritz, and Trevor Darrell.
\newblock Adapting visual category models to new domains.
\newblock In Kostas Daniilidis, Petros Maragos, and Nikos Paragios (eds.),
  \emph{Computer Vision -- ECCV 2010}, pp.\  213--226, Berlin, Heidelberg,
  2010. Springer Berlin Heidelberg.
\newblock ISBN 978-3-642-15561-1.

\bibitem[Sankaranarayanan et~al.(2017)Sankaranarayanan, Balaji, Castillo, and
  Chellappa]{Sankarayaranan17}
Swami Sankaranarayanan, Yogesh Balaji, Carlos~D. Castillo, and Rama Chellappa.
\newblock Generate to adapt: Aligning domains using generative adversarial
  networks.
\newblock \emph{CoRR}, abs/1704.01705, 2017.
\newblock URL \url{http://arxiv.org/abs/1704.01705}.

\bibitem[Schneider et~al.(2012)Schneider, Rasband, and Eliceiri]{pmid22930834}
C.~A. Schneider, W.~S. Rasband, and K.~W. Eliceiri.
\newblock {{N}{I}{H} {I}mage to {I}mage{J}: 25 years of image analysis}.
\newblock \emph{Nat. Methods}, 9\penalty0 (7):\penalty0 671--675, Jul 2012.

\bibitem[Schweikert et~al.(2008)Schweikert, Widmer, Sch\"{o}lkopf, and
  R\"{a}tsch]{Schweikert:2008:EAD:2981780.2981959}
Gabriele Schweikert, Christian Widmer, Bernhard Sch\"{o}lkopf, and Gunnar
  R\"{a}tsch.
\newblock An empirical analysis of domain adaptation algorithms for genomic
  sequence analysis.
\newblock In \emph{Proceedings of the 21st International Conference on Neural
  Information Processing Systems}, NIPS'08, pp.\  1433--1440, USA, 2008. Curran
  Associates Inc.
\newblock ISBN 978-1-6056-0-949-2.
\newblock URL \url{http://dl.acm.org/citation.cfm?id=2981780.2981959}.

\bibitem[Shimodaira(2000)]{Shimodaira00}
Hidetoshi Shimodaira.
\newblock Improving predictive inference under covariate shift by weighting the
  log-likelihood function.
\newblock \emph{Journal of Statistical Planning and Inference}, 90\penalty0
  (2):\penalty0 227--244, 2000.

\bibitem[Shu et~al.(2018)Shu, Bui, Narui, and Ermon]{Shu18}
Rui Shu, Hung~H. Bui, Hirokazu Narui, and Stefano Ermon.
\newblock A {DIRT-T} approach to unsupervised domain adaptation.
\newblock In \emph{Proceedings of the 6th International Conference on Learning
  Representations (ICLR)}, 2018.

\bibitem[Sigal et~al.(2006)Sigal, Milo, Cohen, Geva-Zatorsky, Klein, Alaluf,
  Swerdlin, Perzov, Danon, Liron, Raveh, Carpenter, Lahav, and
  Alon]{pmid16791210}
A.~Sigal, R.~Milo, A.~Cohen, N.~Geva-Zatorsky, Y.~Klein, I.~Alaluf,
  N.~Swerdlin, N.~Perzov, T.~Danon, Y.~Liron, T.~Raveh, A.~E. Carpenter,
  G.~Lahav, and U.~Alon.
\newblock {{D}ynamic proteomics in individual human cells uncovers widespread
  cell-cycle dependence of nuclear proteins}.
\newblock \emph{Nat. Methods}, 3\penalty0 (7):\penalty0 525--531, Jul 2006.

\bibitem[Simonyan \& Zisserman(2014)Simonyan and Zisserman]{vgg16}
Karen Simonyan and Andrew Zisserman.
\newblock Very deep convolutional networks for large-scale image recognition.
\newblock \emph{CoRR}, abs/1409.1556, 2014.
\newblock URL \url{http://arxiv.org/abs/1409.1556}.

\bibitem[Stallkamp et~al.(2012)Stallkamp, Schlipsing, Salmen, and Igel]{gtsrb}
J.~Stallkamp, M.~Schlipsing, J.~Salmen, and C.~Igel.
\newblock Man vs. computer: Benchmarking machine learning algorithms for
  traffic sign recognition.
\newblock \emph{Neural Networks}, pp.\ ~--, 2012.
\newblock ISSN 0893-6080.
\newblock \doi{10.1016/j.neunet.2012.02.016}.
\newblock URL
  \url{http://www.sciencedirect.com/science/article/pii/S0893608012000457}.

\bibitem[Stoeger et~al.(2015)Stoeger, Battich, Herrmann, Yakimovich, and
  Pelkmans]{pmid26014038}
T.~Stoeger, N.~Battich, M.~D. Herrmann, Y.~Yakimovich, and L.~Pelkmans.
\newblock {{C}omputer vision for image-based transcriptomics}.
\newblock \emph{Methods}, 85:\penalty0 44--53, Sep 2015.

\bibitem[Sun \& Saenko(2016)Sun and Saenko]{deep_coral}
Baochen Sun and Kate Saenko.
\newblock Deep coral: Correlation alignment for deep domain adaptation.
\newblock In Gang Hua and Herv{\'e} J{\'e}gou (eds.), \emph{Computer Vision --
  ECCV 2016 Workshops}, pp.\  443--450, Cham, 2016. Springer International
  Publishing.
\newblock ISBN 978-3-319-49409-8.

\bibitem[Sun et~al.(2016)Sun, Feng, and Saenko]{coral}
Baochen Sun, Jiashi Feng, and Kate Saenko.
\newblock Return of frustratingly easy domain adaptation.
\newblock In \emph{Proceedings of the 29th AAAI Conference on Artificial
  Intelligence}, AAAI, 2016.
\newblock URL \url{http://arxiv.org/abs/1511.05547}.

\bibitem[Taigman et~al.(2016)Taigman, Polyak, and Wolf]{Taigman16}
Yaniv Taigman, Adam Polyak, and Lior Wolf.
\newblock Unsupervised cross-domain image generation.
\newblock \emph{CoRR}, abs/1611.02200, 2016.
\newblock URL \url{http://arxiv.org/abs/1611.02200}.

\bibitem[Tzeng et~al.(2015)Tzeng, Hoffman, Darrell, and Saenko]{7410820}
E.~Tzeng, J.~Hoffman, T.~Darrell, and K.~Saenko.
\newblock Simultaneous deep transfer across domains and tasks.
\newblock In \emph{2015 IEEE International Conference on Computer Vision
  (ICCV)}, pp.\  4068--4076, Dec 2015.
\newblock \doi{10.1109/ICCV.2015.463}.

\bibitem[Tzeng et~al.(2014)Tzeng, Hoffman, Zhang, Saenko, and
  Darrell]{DBLP:journals/corr/TzengHZSD14}
Eric Tzeng, Judy Hoffman, Ning Zhang, Kate Saenko, and Trevor Darrell.
\newblock Deep domain confusion: Maximizing for domain invariance.
\newblock \emph{CoRR}, abs/1412.3474, 2014.
\newblock URL \url{http://arxiv.org/abs/1412.3474}.

\bibitem[Tzeng et~al.(2017)Tzeng, Hoffman, Saenko, and
  Darrell]{DBLP:journals/corr/TzengHSD17}
Eric Tzeng, Judy Hoffman, Kate Saenko, and Trevor Darrell.
\newblock Adversarial discriminative domain adaptation.
\newblock \emph{CoRR}, abs/1702.05464, 2017.
\newblock URL \url{http://arxiv.org/abs/1702.05464}.

\bibitem[Vallania et~al.(2017)Vallania, Tam, Lofgren, Schaffert, Azad, Bongen,
  Alsup, Alonso, Davis, Engleman, and Khatri]{Vallania206466}
Francesco Vallania, Andrew Tam, Shane Lofgren, Steven Schaffert, Tej~D. Azad,
  Erika Bongen, Meia Alsup, Michael Alonso, Mark Davis, Edgar Engleman, and
  Purvesh Khatri.
\newblock Leveraging heterogeneity across multiple data sets increases accuracy
  of cell-mixture deconvolution and reduces biological and technical biases.
\newblock \emph{bioRxiv}, 2017.
\newblock \doi{10.1101/206466}.
\newblock URL \url{https://www.biorxiv.org/content/early/2017/10/20/206466}.

\bibitem[van~der Maaten \& Hinton(2008)van~der Maaten and Hinton]{tsne}
Laurens van~der Maaten and Geoffrey Hinton.
\newblock Visualizing data using {t-SNE}.
\newblock \emph{Journal of Machine Learning Research}, 9:\penalty0 2579--2605,
  2008.
\newblock URL \url{http://www.jmlr.org/papers/v9/vandermaaten08a.html}.

\bibitem[{van Opbroek} et~al.(2015){van Opbroek}, Ikram, Vernooij, and {de
  Bruijne}]{opbroek2015}
Annegreet {van Opbroek}, {M. Arfan} Ikram, {Meike W.} Vernooij, and Marleen {de
  Bruijne}.
\newblock Transfer learning improves supervised image segmentation across
  imaging protocols.
\newblock \emph{I E E E Transactions on Medical Imaging}, 34\penalty0
  (5):\penalty0 1018--1030, 2015.
\newblock ISSN 0278-0062.
\newblock \doi{10.1109/TMI.2014.2366792}.

\bibitem[Vapnik(1998)]{Vapnik98}
V.~N. Vapnik.
\newblock \emph{Statistical Learning Theory}.
\newblock Wiley, 1998.

\bibitem[Xiao et~al.(2016)Xiao, Li, Ouyang, and Wang]{Xiao2016}
Tong Xiao, Hongsheng Li, Wanli Ouyang, and Xiaogang Wang.
\newblock Learning deep feature representations with domain guided dropout for
  person re-identification.
\newblock In \emph{Proceedings of the **th Conference on Computer Vision and
  Pattern Recognition}, CVPR'16, 2016.
\newblock URL \url{http://arxiv.org/abs/1604.07528}.

\bibitem[Xu \& Yang(2011)Xu and Yang]{Xu:2011fj}
Qian Xu and Qiang Yang.
\newblock {A Survey of Transfer and Multitask Learning in Bioinformatics}.
\newblock \emph{Journal of Computing Science and Engineering}, August 2011.

\bibitem[Yang \& Hospedales(2015)Yang and Hospedales]{Yang2015}
Yongxin Yang and Timothy~M. Hospedales.
\newblock A unified perspective on multi-domain and multi-task learning.
\newblock In \emph{Proceedings of the 3d International Conference on
  Representation Learning}, ICLR'15, 2015.

\bibitem[Yi et~al.(2017)Yi, Zhang, Tan, and Gong]{Yi17}
Zili Yi, Hao~(Richard) Zhang, Ping Tan, and Minglun Gong.
\newblock Dualgan: Unsupervised dual learning for image-to-image translation.
\newblock In \emph{{ICCV}}, pp.\  2868--2876. {IEEE} Computer Society, 2017.

\bibitem[Zhang et~al.(2012)Zhang, Zhang, and Ye]{NIPS2012_4684}
Chao Zhang, Lei Zhang, and Jieping Ye.
\newblock Generalization bounds for domain adaptation.
\newblock In F.~Pereira, C.~J.~C. Burges, L.~Bottou, and K.~Q. Weinberger
  (eds.), \emph{Advances in Neural Information Processing Systems 25}, pp.\
  3320--3328. Curran Associates, Inc., 2012.
\newblock URL
  \url{http://papers.nips.cc/paper/4684-generalization-bounds-for-domain-adaptation.pdf}.

\bibitem[Zhang et~al.(2018)Zhang, Ding, Li, and Ogunbona]{partialDA2}
Jing Zhang, Zewei Ding, Wanqing Li, and Philip Ogunbona.
\newblock Importance weighted adversarial nets for partial domain adaptation.
\newblock \emph{CoRR}, abs/1803.09210, 2018.
\newblock URL \url{http://arxiv.org/abs/1803.09210}.

\bibitem[Zhu et~al.(2017)Zhu, Park, Isola, and Efros]{Zhu17}
Jun{-}Yan Zhu, Taesung Park, Phillip Isola, and Alexei~A. Efros.
\newblock Unpaired image-to-image translation using cycle-consistent
  adversarial networks.
\newblock In \emph{{ICCV}}, pp.\  2242--2251. {IEEE} Computer Society, 2017.

\end{thebibliography}

\clearpage
\appendix
\section{Extended state-of-the-art: image translation}
\label{sec:extended_soa}
In the field of computer vision, another way of mapping examples in one domain onto the other domain is image-to-image translation. In the supervised case (the true pairs made of an image and its translation are given), Pic2Pix \citep{Isola16} trains a conditional GAN to discriminate true pairs from fake ones. In the unsupervised case, another loss is designed to enforce cycle consistency (simultaneously learning the mapping $\phi$ from domain $A$ to $B$, $\psi$ from $B$ to $A$, and requiring $\phi o \psi = $Id) \citep{Zhu17,Yi17}. Note that translation approaches do not {\em per se} address domain adaptation as they are agnostic w.r.t. the classes. Additional losses are used to overcome this limitation:
Domain transfer network (DTN) \citep{Taigman16} uses an auto-encoder-like loss in the latent space; GenToAdapt \citep{Sankarayaranan17} uses a classifier loss in the latent space; UNIT \citep{Liu17} uses a VAE loss.

StarGAN \citep{Choi17} combines image-to-image translation with a GAN, where the discriminator is trained to discriminate true from fake pairs on the one hand, and the domain on the other hand. ComboGAN \citep{Anoosheh17} learns two networks per domain, an encoder and a decoder. DIRT-T \citep{Shu18} uses a conditional GAN and a classifier in the latent space, with two additional losses, respectively enforcing the cluster assumption (the classifier boundary should not cross high density region) and a virtual adversarial training (the hypothesis should be invariant under slight perturbations of the input). \\
Interestingly, DA and MDL (like deep learning in general) tend to combine quite some losses; two benefits are expected from using a mixture of losses, a smoother optimization landscape and a good stability of the representation \citep{DBLP:journals/jmlr/BousquetE02}. 
\section{Proofs}
\label{proof}
\subsection{Definition of the $\mathcal{H}$-divergence}
\label{sec:def}
\begin{definition}
    \citep{shai2004, shai2006,shai2010} Given a domain $\mathcal X$, two distributions $\mathcal D$ and $\mathcal D'$ over that domain and a binary hypothesis class $\mathcal H$ on $\cal X$, the $\mathcal H$-divergence between $\mathcal D$ and $\mathcal D'$ is defined as:
    \begin{equation*}
    d_{\mathcal H} (\mathcal D, \mathcal D') = 2~ \underset{h \in \mathcal H}{\sup} \vert \mathbb P_{\mathcal D}(h(\mathbf{x})=1) - \mathbb P_{\mathcal D'}(h(\mathbf{x})=1) \vert
    \end{equation*}
\end{definition}

\subsection{Bounding MDL loss using the $\mathcal{H}$-divergence}
\label{sec:thm1}
\begin{thm}
\label{theorem1}
    Given an input space $\mathcal X$, we consider $n$ distributions $\mathcal D_i$ over $\mathcal X \times \{0;1\}$ and a hypothesis class $\mathcal H$ on $\mathcal X$ of VC dimension $d$. 
    Let $\alpha$ and $\gamma$ be in the simplex of dimension $n$. If $S$ is a sample of size $m$ which contains $\gamma_i m$ samples from $\mathcal D_i$, and $\hat{h}$ is the empirical minimizer of $\sum_i \alpha_i \hat{\epsilon}_i$ on $(S_i)_i$, then for any $\delta >0$, with probability at least $1-\delta$, the compound empirical error is upper bounded as:
    \begin{equation}
        \sum_i \epsilon_i(\hat{h}) \leq \sum_i \epsilon_i^\star + 4n B(\alpha) + 2\sum_{i \le j} (\alpha_i + \alpha_j )\left( d_{\mathcal{H}} (\mathcal D_i^X, \mathcal{D}_j^X) +\beta_{i,j}\right)
    \end{equation}
    with \[B(\alpha) = \sqrt{\sum_j \dfrac{\alpha_j^2}{\gamma_j}} \sqrt{\dfrac{2d\log(2(m+1)) +\log(\tfrac{4}{\delta})}{m}}\]
    and \[
    \beta_{i,j} = \underset{h \in \mathcal H}{min}~ \left( \epsilon_i(h) + \epsilon_j(h)\right)
    \]
\end{thm}

A tighter bound can be obtained by replacing $d_{\mathcal{H}} (\mathcal D_i, \mathcal{D}_j)$ with $\tfrac{1}{2}d_{\mathcal{H}\Delta\mathcal{H}} (\mathcal D_i, \mathcal{D}_j)$. The $\mathcal{H}\Delta\mathcal{H}$-divergence \citep{shai2010} operates on the symmetric difference hypothesis space $\mathcal{H}\Delta\mathcal{H}$. However,  divergence $\mathcal{H}\Delta\mathcal{H}$ does not lend itself to empirical estimation: even~\citet{shai2010} fall back on $\mathcal{H}$-divergence in their empirical validation.

\paragraph{Proof of theorem 2} For $i, j$ we note $\beta_{i,j} = \epsilon_i(h_{i,j}^\star) + \epsilon_j(h_{i,j}^\star) = \underset{h \in \mathcal H}{min}~ (\epsilon_i(h) + \epsilon_j(h))$. For $\alpha$ in the $n$-dimensional simplex and $h \in \mathcal H$, we note $\epsilon_\alpha(h) = \sum_i \alpha_i \epsilon_i({h})$. 

We have for $\alpha$ in the simplex of dimension $n$, $h \in \mathcal H$ and $j \in \{1,\ldots,m\}$, using the triangle inequality (similarly to the proof of Theorem 4 in~\citep{shai2010})

\begin{align*}
    \vert \epsilon_\alpha(h) - \epsilon_j(h) \vert = & \left| \sum_i \alpha_i \left( \mathbb E_{ \mathbf{x},y\sim \mathcal D_i}\vert h(\mathbf{x}) - y\vert - \mathbb E_{ \mathbf{x},y\sim \mathcal D_j}\vert h(\mathbf{x}) - y\vert\right) \right| \\
     \leq & \sum_i \alpha_i \left| \mathbb E_{ \mathbf{x},y\sim \mathcal D_i}\vert h(\mathbf{x}) - y\vert - \mathbb E_{ \mathbf{x},y\sim \mathcal D_j}\vert h(\mathbf{x}) - y\vert \right| \\
     \leq & \sum_i \alpha_i 
     \left| \mathbb E_{ \mathbf{x},y\sim \mathcal D_i}\vert h(\mathbf{x}) - y\vert - \mathbb E_{ \mathbf{x}\sim \mathcal D_i}\vert h(\mathbf{x}) - h_{i,j}^\star(\mathbf{x})\vert \right|\\
&     + \alpha_i~ \left| \mathbb E_{ \mathbf{x}\sim \mathcal D_i}\vert h(\mathbf{x}) - h_{i,j}^\star(\mathbf{x})\vert - \mathbb E_{ \mathbf{x}\sim \mathcal D_j}\vert h(\mathbf{x}) - h_{i,j}^\star(\mathbf{x})\vert \right| \\
&     + \alpha_i~ \left| \mathbb E_{ \mathbf{x}\sim \mathcal D_j}\vert h(\mathbf{x}) - h_{i,j}^\star(\mathbf{x})\vert - \mathbb E_{ \mathbf{x},y\sim \mathcal D_j}\vert h(\mathbf{x}) - y\vert \right| 
     \\
     \leq & \sum_i \alpha_i \left( \beta_{i,j} + d_{\mathcal{H}}(\mathcal{D}_i, \mathcal{D}_j) \right)
\end{align*}

The last line follows from the definitions of $\beta_{i,j}$ and $\mathcal{H}$-divergence. Thus using lemma 6 in~\citep{shai2010}
\begin{align*}
    \epsilon_j(\Hat{h}) \leq & \epsilon_\alpha(\Hat{h}) + \sum_i \alpha_i \left( \beta_{i,j} + d_{\mathcal{H}}(\mathcal{D}_i, \mathcal{D}_j) \right) \\
    \leq & \hat{\epsilon}_\alpha(\Hat{h}) + 2B(\alpha) +\sum_i \alpha_i \left( \beta_{i,j} + d_{\mathcal{H}}(\mathcal{D}_i, \mathcal{D}_j) \right) \\
    \leq & \hat{\epsilon}_\alpha(h_{j}^\star) + 2B(\alpha) +\sum_i \alpha_i \left( \beta_{i,j} + d_{\mathcal{H}}(\mathcal{D}_i, \mathcal{D}_j) \right) \\
    \leq & {\epsilon}_\alpha(h_{j}^\star) + 4B(\alpha) +\sum_i \alpha_i \left( \beta_{i,j} + d_{\mathcal{H}}(\mathcal{D}_i, \mathcal{D}_j) \right) \\
    \leq & \epsilon_i^\star + 4B(\alpha) +2\sum_i \alpha_i \left( \beta_{i,j} + d_{\mathcal{H}}(\mathcal{D}_i, \mathcal{D}_j) \right) \\
\end{align*}
with \[B(\alpha) = \sqrt{\sum_j \dfrac{\alpha_j^2}{\beta_j}} \sqrt{\dfrac{2d\log(2(m+1)) +\log(\tfrac{4}{\delta})}{m}}\] Hence the result.$\Box$
\subsection{Bounding domain imbalance}
\label{sec:proofProp2}
\paragraph{Proof of proposition 1} We have for $h \in \mathcal H$ and $j \in [1,\ldots,m]$,  using the triangle inequality and the definition of $\epsilon_i^\star$ (similarly to the proof of Theorem 1 in~\citep{shai2006})
\begin{align*}
\epsilon_j(h) &= \mathbb P_{ \mathbf{x},y\sim \mathcal D_j}(h(\mathbf{x}) \neq y)\\
&= \mathbb E_{ \mathbf{x},y\sim \mathcal D_j}\vert h(\mathbf{x}) - y\vert\\
&\leq\mathbb E_{\mathcal D_j^X}\vert h(\mathbf{x}) - h_j^\star(\mathbf{x})\vert +\mathbb E_{\mathcal D_j}\vert h_j^\star(\mathbf{x}) - y\vert \\
&\leq\mathbb E_{\mathcal D_j^X}\vert h(\mathbf{x}) - \dfrac{1}{n} \sum_i h_i^\star (\mathbf{x}) \vert 
+\mathbb E_{\mathcal D_j^X}\vert \dfrac{1}{n} \sum_i h_i^\star (\mathbf{x}) - h_j^\star(\mathbf{x})\vert + \epsilon_j^\star\\
&\leq \dfrac{1}{n} \sum_i\mathbb E_{\mathcal D_j^X}\vert h(\mathbf{x}) -  h_i^\star (\mathbf{x}) \vert 
+ \dfrac{1}{n} \sum_i\mathbb E_{\mathcal D_j^X}\vert h_i^\star (\mathbf{x}) - h_j^\star(\mathbf{x})\vert + \epsilon_j^\star\\
\end{align*}
We have for $i$
\begin{align*}
\mathbb E_{\mathcal D_j^X}\vert h(\mathbf{x}) -  h_i^\star (\mathbf{x}) \vert & \leq 
\mathbb E_{\mathcal D_i^X}\vert h(\mathbf{x}) - h_i^\star(\mathbf{x})\vert +
\vert\mathbb E_{\mathcal D_i^X}\vert h(\mathbf{x}) - h_i^\star(\mathbf{x})\vert -\mathbb E_{\mathcal D_j^X}\vert h(\mathbf{x}) - h_i^\star(\mathbf{x})\vert \vert \\
&\leq \epsilon_i(h) + \epsilon_i^\star + d_{\mathcal H} (\mathcal D_i^X, \mathcal D_j^X)
\end{align*}
The second line follows from the triangle inequality and the definition of the $\mathcal{H}$-divergence. Thus
\begin{equation}
    \epsilon_j(h) \leq \dfrac{1}{n} \sum_i \left( \epsilon_i(h) + \epsilon_i^\star + d_{\mathcal H} (\mathcal D_i^X, \mathcal D_j^X) +\mathbb E_{\mathcal D_j^X}\vert h_i^\star (\mathbf{x}) - h_j^\star(\mathbf{x})\vert \right) + \epsilon_j^\star
\end{equation}

By symmetry we obtain
\begin{align*}
    \dfrac{1}{n} \sum_i \epsilon_i(h) &\leq \epsilon_j(h) + \dfrac{1}{n} \sum_i \left( \epsilon_i^\star + d_{\mathcal H} (\mathcal D_i^X, \mathcal D_j^X) +\mathbb E_{\mathcal D_i^X}\vert h_i^\star (\mathbf{x}) - h_j^\star(\mathbf{x})\vert \right) + \epsilon_j^\star
\end{align*}
Thus the result.$\Box$
\begin{prop}\label{prop2} Given a domain $\mathcal X$, $m$ distributions $\mathcal D_i$ over $\mathcal X \times \{0;1\}$ and a hypothesis class $\mathcal H$ on $\mathcal X$, we have for $h \in \mathcal H$ and $j \in [1,\ldots,m]$
\begin{equation}
\vert \epsilon_j(h) - \dfrac{1}{n} \underset{i}{\sum} \epsilon_i(h)\vert \leq 2\left( \epsilon_j^\star + \dfrac{1}{n} \underset{i}{\sum} \epsilon_i^\star\right) + \epsilon_j(h^\star) + \beta + \dfrac{1}{n} \underset{i}{\sum} d_{\mathcal H} (\mathcal D_i^X, \mathcal D_j^X) + \tfrac{1}{2} d_{\mathcal{H}\Delta\mathcal{H}} (\mathcal D_i, \mathcal{D}_j)
\end{equation}
where 
\[
\beta = \underset{j}{\sum} \epsilon_j(h^\star) = \underset{h \in \mathcal H}{min}~ \underset{j}{\sum} \epsilon_j(h)
\]
\end{prop}
\paragraph{Proof} For $i,j \in [i, \ldots, m]$ we have
\begin{align*}
   \mathbb E_{\mathcal D_i}\vert h_i^\star(\mathbf{x}) - h_j^\star(\mathbf{x})\vert &\leq\mathbb E_{\mathcal D_i}\vert h_j^\star(\mathbf{x}) - h^\star(\mathbf{x})\vert + 
   \mathbb E_{\mathcal D_i}\vert h^\star(\mathbf{x}) - h_i^\star(\mathbf{x})\vert\\
    &\leq\mathbb E_{\mathcal D_j}\vert h_j^\star(\mathbf{x}) - h^\star(\mathbf{x})\vert + \tfrac{1}{2} d_{\mathcal{H}\Delta\mathcal{H}} (\mathcal D_i, \mathcal{D}_j) + \epsilon_i(h^\star) + \epsilon_i^\star\\
    &\leq \epsilon_i(h^\star) + \epsilon_j(h^\star) + \epsilon_i^\star + \epsilon_j^\star + \tfrac{1}{2} d_{\mathcal{H}\Delta\mathcal{H}} (\mathcal D_i, \mathcal{D}_j) 
\end{align*}
The second line follows from Lemma 3 from~\citep{shai2010}, and the third from the triangle inequality. From this and proposition~\ref{prop1} we obtain the result.
$\Box$
\paragraph{Corollaries for the 2-domain case}
\begin{cor}\label{coro3}
    Given a domain $\mathcal X$, two distributions $\mathcal D_S$ and $\mathcal D_T$ over $\mathcal X \times \{0,1\}$ and a hypothesis class $\mathcal H$ on $\mathcal X$, we have for $h \in \mathcal H$
    \begin{equation}
    \vert \epsilon_S(h) - \epsilon_T(h)\vert \leq \epsilon_T^\star +\epsilon_S^\star + \Delta + d_{\mathcal H} (\mathcal D_S^X, \mathcal D_T^X)
    \end{equation}
    with $\Delta = \max(E_{\mathcal D_T^X}\vert h_S^\star(\mathbf{x}) - h_T^\star(\mathbf{x})\vert, E_{\mathcal D_S^X}\vert h_S^\star(\mathbf{x}) - h_T^\star(\mathbf{x})\vert) $
\end{cor}

\begin{cor}\label{coro4}
Given a domain $\mathcal X$, two distributions $\mathcal D_S$ and $\mathcal D_T$ over $\mathcal X \times \{0;1\}$ and a hypothesis class $\mathcal H$ on $\mathcal X$, we have for $h \in \mathcal H$
\begin{equation}
\vert \epsilon_S(h) - \epsilon_T(h)\vert \leq 2(\epsilon_T^\star +\epsilon_S^\star) + \beta + \tfrac{1}{2} d_{\mathcal{H}\Delta\mathcal{H}} (\mathcal D_S, \mathcal{D}_T) + d_{\mathcal H} (\mathcal D_S^X, \mathcal D_T^X)
\end{equation}
where 
\[
\beta = \epsilon_S(h^\star) + \epsilon_T(h^\star) = \underset{h \in \mathcal H}{min}~ \epsilon_S(h) + \epsilon_T(h)
\]
\end{cor}

\section{\pheno\ dataset}
\label{biodata}
\begin{figure}
\centerline{\includegraphics[width=1.1\linewidth]{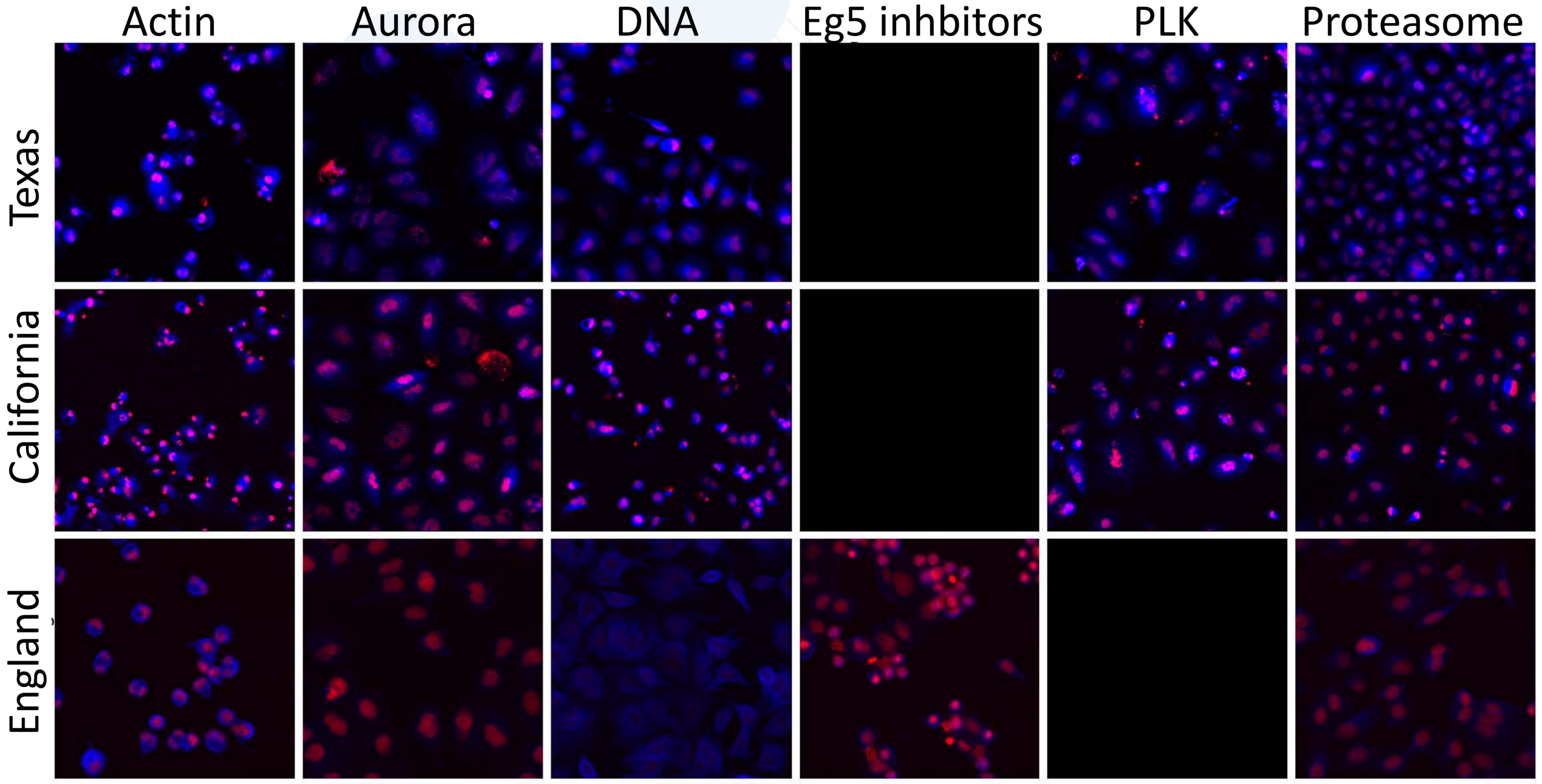}}
\caption{Examples from six classes in the Bio dataset (red: cell nuclei, blue: cell cytoplasm, magnification: 10X). Empty squares: the domain does not contain any known examples from this class. Best seen in color.}
\label{fig:image_bio}
\end{figure}

\subsection{Texas domain}
This dataset is extracted from that published in~\citep{pmid26655497}. It contains 455 biologically active images, in 11 classes, on four 384-well plates, in three channels: H2B-CFP, XRCC5-YFP and cytoplasmic-mCherry. 
Our analysis used 10 classes:  'Actin', 'Aurora', 'DNA', 'ER', 'HDAC', 'Hsp90', 'MT', 'PLK', 'Proteasome', 'mTOR'.

On top of the quality control from the original paper, a visual quality control was implemented to remove images with only apoptotic cells, and XRCC5-YFP channel images were smoothed using a median filter of size 2 using SciPy~\citep{scipy}.

\subsection{California domain}
This dataset is designed to be similar to the Texas domain~\citep{pmid26655497}, generated using the same cell line, but in a different laboratory, by a different biologist, and using different equipment. It contains 1,077 biologically active images, in 10 classes, on ten 384-well plates, in three channels: H2B-CFP, XRCC5-YFP and cytoplasmic-mCherry. The classes are: 'Actin', 'Aurora', 'DNA', 'ER', 'HDAC', 'Hsp90', 'MT', 'PLK', 'Proteasome', 'mTOR'.

\paragraph{Cell culture, drug screening and image acquisition}
Previously~\citep{pmid26655497}, retroviral transduction of a marker plasmid "pSeg" was used to stably express H2B-CFP and cytoplasmic-mCherry tags in A549 human lung adenocarcinoma cells. A CD-tagging approach~\citep{pmid16791210} was used to add an N-terminal YFP tag to endogenous XRCC5.

Cells were maintained in RPMI1640 media containing 10\% FBS, 2 mM glutamine, 50 units/ml penicillin, and 50 $\mu$g/ml streptomycin (all from Life Technologies, Inc.), at $37^\circ C$, 5\% CO$^2$ and 100\% humidity. 
24h prior to drug addition, cells were seeded onto 384-well plate at a density of 
1200 cells/well. Following compound addition, cells were incubated at $37^\circ C$ for 48 hours. Images were then acquired using a 
GE InCell Analyzer 2000. One image was acquired per well using a 10x objective lens with 2x2 binning. 

\paragraph{Image processing}
Uneven illumination was corrected as described in~\citep{pmid26014038}. Background noise was removed using the ImageJ RollingBall plugin~\citep{pmid22930834}. Images were segmented, object features extracted and biological activity determined as previously described~\citep{pmid26655497}. A visual quality control was implemented to remove images with obvious anomalies (e.g. presence of a hair or out-of-focus image) and images with only apoptotic cells. YFP-XRCC5 channel images were smoothed using a median filter of size 2.

\subsection{England domain}
This dataset was published by~\citet{pmid20530715} and retrieved from~\citep{pmid22743765}. It contains 879 biologically active images of  MCF7 breast adenocarcinoma cells, in 15 classes on 55 96-well plates, in 3 channels: Alexa Fluor 488 (Tubulin), Alexa Fluor 568 (Actin) and DAPI (nuclei). Classes with fewer than 15 images and absent from the other datasets ("Calcium regulation", "Cholesterol", "Epithelial", 
"MEK", "mTOR") were not used, which leaves 10 classes: 'Actin', 'Aurora', 'DNA', 'ER', 'Eg5 inhibitor', 'HDAC', 'Kinase', 'MT', 'Proteasome', 'Protein synthesis'.

\paragraph{Image processing}
As the images were acquired using a 20X objective, they were stitched using ImageJ plugin~\citep{pmid19346324} and down-scaled 2 times. Cells thus appear the same size as in the other domains. Images were segmented, object features extracted and biological activity obtained as previously described~\citep{pmid26655497}. A visual quality control was implemented to remove images with obvious anomalies and images with only apoptotic cells. Images with too few cells were also removed: an Otsu filter~\citep{otsu} was used to estimate the percentage of pixels containing nuclei in each image, and images with less than 1\% nuclear pixels were removed. Tubulin channel images were smoothed using a median filter of size 2.

\subsection{Common image pre-processing}
\label{sec:bio_images}
Images which were not significantly distinct from negative controls were identified as previously~\citep{pmid26655497} and excluded from our analysis. Previous work on the England dataset further focused on images which "clearly [have] one of 12 different primary mechanims of action"~\citep{pmid22743765}. We chose not to do so, since it results in a simpler problem (90\% accuracy easy to reach) with much less room for improvement.

Images from all domains were down-scaled 4 times and flattened to form RGB images. Images were normalized by subtracting the intensity values from negative controls (DMSO) of the same plate in each channel. England, Texas and California share images for cell nucleus and cytoplasm, but their third channel differs: Texas and California shows the protein XRCC5, whereas England shows the Actin protein. Therefore, the experiments which combine Texas and England, and California and England used only the first two channels, feeding an empty third channel into the network. Similarly, profiles contain 443 features which are related to the first two channels, and 202 features which are related to the third channel. Only the former were used in experiments which involve the England dataset. 

\subsection{Semi-supervised MDL experiments}
\label{sec:bio_classes}
\begin{table}[!h]
    \centering
    \begin{tabular}{c|ll}
        \toprule
        Shift & Dom. 2, labeled classes & Domain 2, unlabeled classes  \\
        \midrule
        E-C & HDAC, Proteasome, Actin, Aurora & DNA, MT, ER\\
        C-T & DNA, HDAC, MT,
        ER, Aurora,
        mTOR, PLK & Actin, Proteasome, Hsp90 \\
        T-E & DNA, MT, Proteasome, Actin, ER& Aurora, HDAC, Actin \\
        C-T-E & DNA, MT, Proteasome, Actin, ER& Aurora, HDAC, Actin \\
        \bottomrule
    \end{tabular}
    \caption{Class content for the \pheno\ experiments in table~\ref{table:bio}. In all cases, the first domain contains the same classes as domain 2, though with labeled examples from all classes. These classes were picked as those with best classification accuracy in an unsupervised setting; results are similar when picking the classes with worst classification accuracy. 10 labeled images per class were used for training.}
    \label{tab:my_label}
\end{table}

\clearpage
\section{Experimental settings}
\subsection{Architecture}
\label{sec:architecture}
As in~\citep{ganin,DBLP:journals/corr/TzengHZSD14}, a bottleneck fully connected layer is added after the last dense layer of VGG-16. Learning rates on weights (resp. biases) from "from scratch" layers is ten (resp, twenty) times that on parameters of fine-tuned layers. Instance normalization is used on \digits, whereas global normalization is used on \office\ and \pheno.
\begin{table}[ht]
    \centering 
    \begin{tabular}{cc}
        \toprule
         \office\ and \pheno & \digits \\
         \toprule
         \multicolumn{2}{c}{Feature extractor} \\
         \midrule
         VGG-16, layers Conv$_1$ to FC$_7$ & 5x5 conv. 32; ReLU; 2x2 max pool, 2x2 stride\\
         Fully connected 256 & 5x5 conv. 48; ReLU; 2x2 max pool, 2x2 stride\\
         \toprule
         \multicolumn{2}{c}{Classifier} \\
         \midrule
         Output of feature extractor& Output of feature extractor\\
          & Fully connected 100; ReLU\\
          & Fully connected 100; ReLU\\
         Fully connected $L$; Softmax& Fully connected $L$; Softmax\\
         \toprule
         \multicolumn{2}{c}{Domain discriminator} \\
         \midrule
         Output of feature extractor & Output of feature extractor \\
         Gradient reversal layer & Gradient reversal layer\\
         Fully connected 1,024; ReLU; Dropout (0.5) & Fully connected 100; ReLU\\
         Fully connected 1,024; ReLU; Dropout (0.5)& \\
         Fully connected $i$; Activation & Fully connected $i$; Activation\\
         \midrule
    \end{tabular}
    \caption{Architectures. In the case when considering only two domains, $i=1$ and the last activation of domain discriminators is a sigmoid. 
    When considering three domains, $i=3$ and the activation is a softmax. Knowledge discriminator architecture is identical to that of domain discriminators 
    without the gradient reversal layer.}
    \label{tab:arch_dd}
\end{table}

\subsection{Hyper-parameter search}
\label{sec:hyperparam}
\begin{table}[!h]
    \centering
    \begin{tabular}{l|c|c}
        \toprule
        Parameter & \digits\ and Signs & \pheno  \\
        \midrule
        Learning rate (lr) & $10^{-3}, 10^{-4}$ &  $10^{-4}$ (+ $10^{-5}$ for 3-dom.)\\
        Individual lr & NA & True, False \\
        Lr schedule & \multicolumn{2}{c}{Exponentially decreasing, constant} \\
        $\lambda$ & \multicolumn{2}{c}{$0.1, 0.8$} \\
        $\lambda$ schedule & \multicolumn{2}{c}{Exponentially increasing, constant} \\
        $\zeta$ &  \multicolumn{2}{c}{$0.1, 0.8$} \\
        \bottomrule
    \end{tabular}
    \caption{Range of hyper-parameters which were evaluated in cross-validation experiments. Exponentially decreasing schedule, exponentially increasing schedule, indiv. lr (learning rates from layers which were trained from scratch are multiplied by 10), as in~\citep{ganin}.}
\end{table}
\clearpage
\section{Additional results}
\subsection{3-domain results on \office}
\begin{table}[!h]
   \caption{Classification results on target test set in the semi-supervised DA setting (average and stdev on 5 seeds or folds)}
   \tabcolsep=0.1cm
   \small
   \begin{center}
   \begin{tabular}{lllll}
     \toprule
     \multicolumn{2}{l}{Sources}& D, W & A, W & A, D  \\
     \multicolumn{2}{l}{Target}& Amazon & DSLR & Webcam  \\
     
    \midrule
    \multicolumn{2}{l}{Baseline} &41.7~(1.0)&90.9~(1.3)&89.4~(1.5)\\
     \multirow{3}{*}{FT} &\DANN &57.5~(1.6)&92.3~(1.8)&91.2~(0.7)\\
    &MADA &37.5~(6.8)&Not conv.&88.3~(0.7)\\
    &\XX &54.5~(3.8)&92.1~(2.6)&92.0~(1.0)\\
\end{tabular}
\end{center}
\end{table}
\clearpage
\subsection{tSNE visualization}
\label{appendix:tsne}
\begin{figure}
\includegraphics[width=\linewidth]{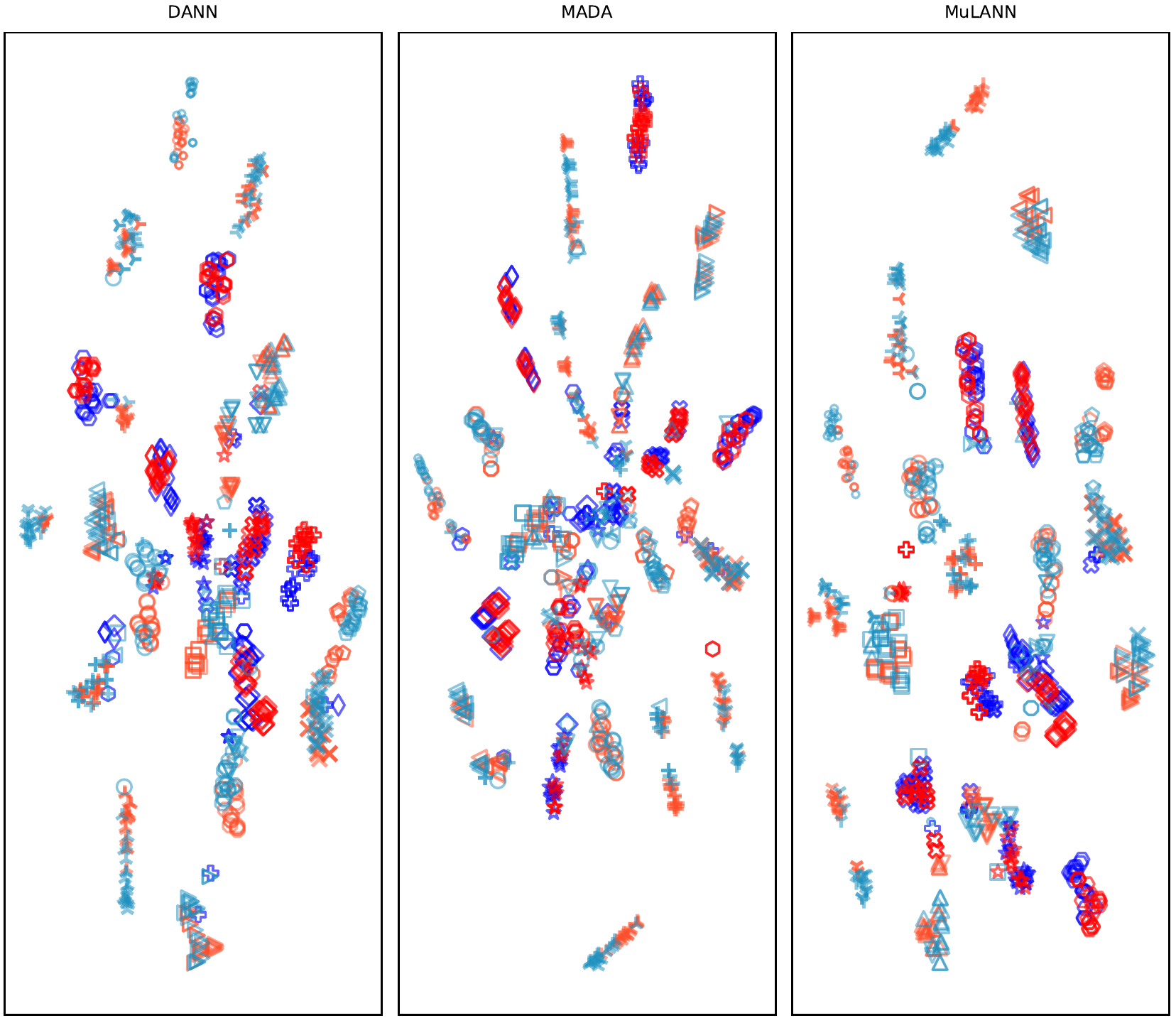}
\caption{Visualization of class features on Webcam (red) > Amazon (blue). Dimmer colors indicate classes for which labeled examples are available in both domains.}
\label{fig:tsne_large}
\end{figure}

We use tSNE~\citep{tsne} to visualize the common feature space in the example of Webcam $\rightarrow$ Amazon. Fig.~\ref{fig:tsne_large} shows that classes are overall better separated with \XX. In particular, when using \XX, unlabeled examples (blue) are both more grouped and closer to labeled points from the other domain.
\clearpage
\subsection{Semi-supervised MDL on the Bio dataset}
  \renewcommand\thetable{5}
 \begin{table}[!h]
    \caption{\pheno\ average test classification results on all domain (average and stdev on 5 folds). P stands for "profiles", "lab." for labeled and "unlab." for unlabeled. Baselines are obtained by training \XX\ with $\lambda=0$ (NN) and LDA+k-NN (P) on both domains. Results were obtained in the non-fully transductive setting, without hyper-parameter optimization.}
  \label{table:bio_non_transductive}
  \centering
   \tabcolsep=0.1cm
   \small
   \begin{tabular}{lllllll|ll}
     \toprule
      Shift & Image set & \# classes & Baseline NN & \DANN & MADA & \XX & {Baseline P} & P+Coral\\
     \midrule
\multirow{3}{*}{E-C}&E&7&74.1~(5.4)&71.6~(5.8)&63.6~(6.1)&72.7~(4.0)&78.1~(8.0)&66.4~(2.4)\\ 
&C lab.&4&98.3~(0.6)&96.1~(1.5)&92.3~(5.2)&89.1~(6.4)&98.2~(2.4)&94.1~(2.3)\\ 
&C unlab.&3&0.4~(0.7)&34.8~(20.7)&14.5~(7.4)&25.7~(12.3)&21.5~(8.4)&36.8~(3.7)\\ 
\midrule 
\multirow{3}{*}{C-T}&C&10&91.4~(1.8)&87.0~(2.2)&87.9~(3.9)&89.3~(1.8)&96.1~(1.1)&93.3~(1.8)\\ 
&T lab.&7&93.7~(1.3)&91.0~(4.4)&86.7~(7.5)&89.2~(1.2)&96.2~(2.4)&92.8~(3.2)\\ 
&T unlab.&3&24.4~(10.0)&61.4~(7.7)&56.2~(14.0)&77.7~(4.0)&59.6~(11.3)&87.6~(8.2)\\ 
\midrule 
\multirow{3}{*}{T-E}&T&7&95.2~(2.2)&90.3~(5.4)&93.7~(3.0)&88.2~(6.4)&94.2~(6.3)&92.6~(4.0)\\ 
&E lab.&4&75.2~(9.7)&61.9~(8.5)&71.0~(12.7)&72.8~(14.2)&81.1~(8.8)&61.2~(4.0)\\ 
&E unlab.&3&5.7~(6.6)&31.4~(17.5)&26.0~(19.4)&51.3~(13.5)&16.1~(9.5)&25.7~(12.6)\\ 
\midrule 
\multirow{3}{*}{C-T-E}&C&7&94.7~(2.0)&91.7~(1.4)&82.7~(3.8)&93.9~(1.7)&94.1~(2.0)&89.4~(2.2)\\ 
&T&7&94.8~(2.1)&93.7~(4.7)&86.5~(4.2)&94.9~(2.1)&97.8~(0.5)&89.6~(8.0)\\ 
&E lab.&4&74.1~(9.8)&67.7~(12.8)&48.2~(11.7)&66.6~(9.0)&74.7~(10.5)&55.6~(7.5)\\ 
&E unlab.&3&10.7~(9.7)&48.6~(21.9)&22.6~(11.3)&69.3~(21.1)&36.3~(6.6)&52.5~(22.5)\\ 
     \bottomrule
   \end{tabular}
 \end{table}
 
\subsection{Impact of $p-p^\star$ on a domain without unlabeled datapoints}
\begin{figure}[!h]
    \centerline{
    \includegraphics[width=0.8\linewidth]{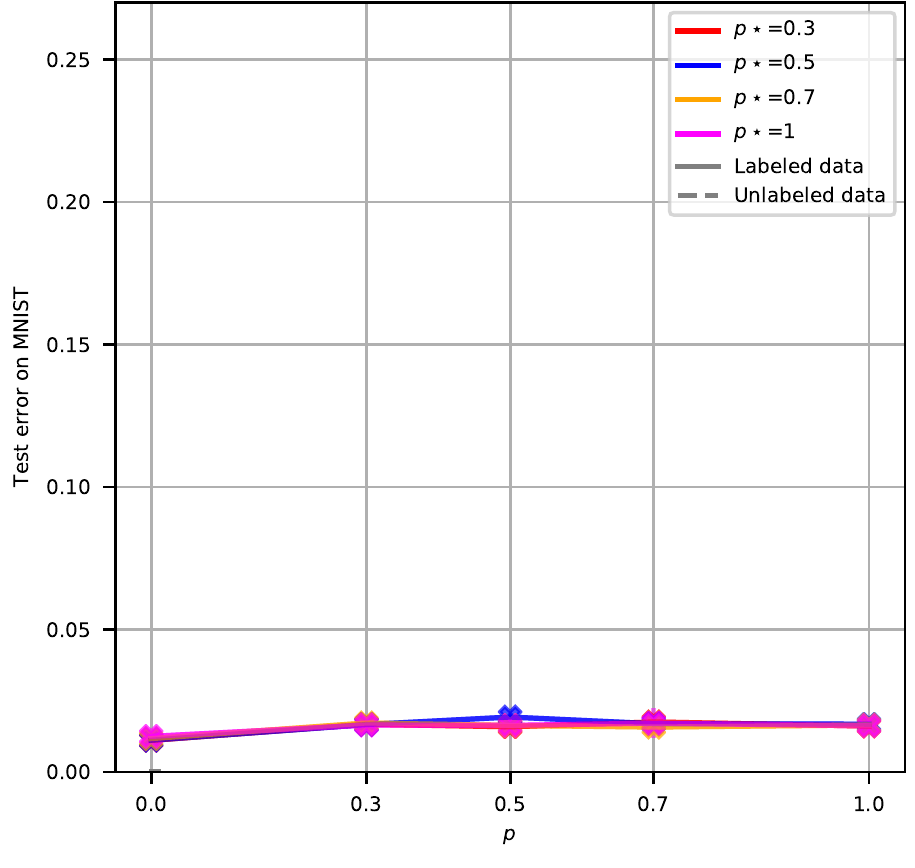}}
    \caption{Impact of parameter $p$ in comparison with $p^\star$  on MNIST $\leftrightarrow$ MNIST-M. $p=0$ corresponds to DANN (see text for details): no data flowed through the KUD module. We can see that different values of $(p, p^\star)$ do not influence the accuracy on a domain which did not have any unlabaled datapoints from extra classes (MNIST in this case).}
    \label{fig:p_eval_source}
\end{figure}

 \subsection{Asymmetry results on \pheno}
 \begin{figure}[!h]
\centerline{\includegraphics[width=0.8\linewidth]{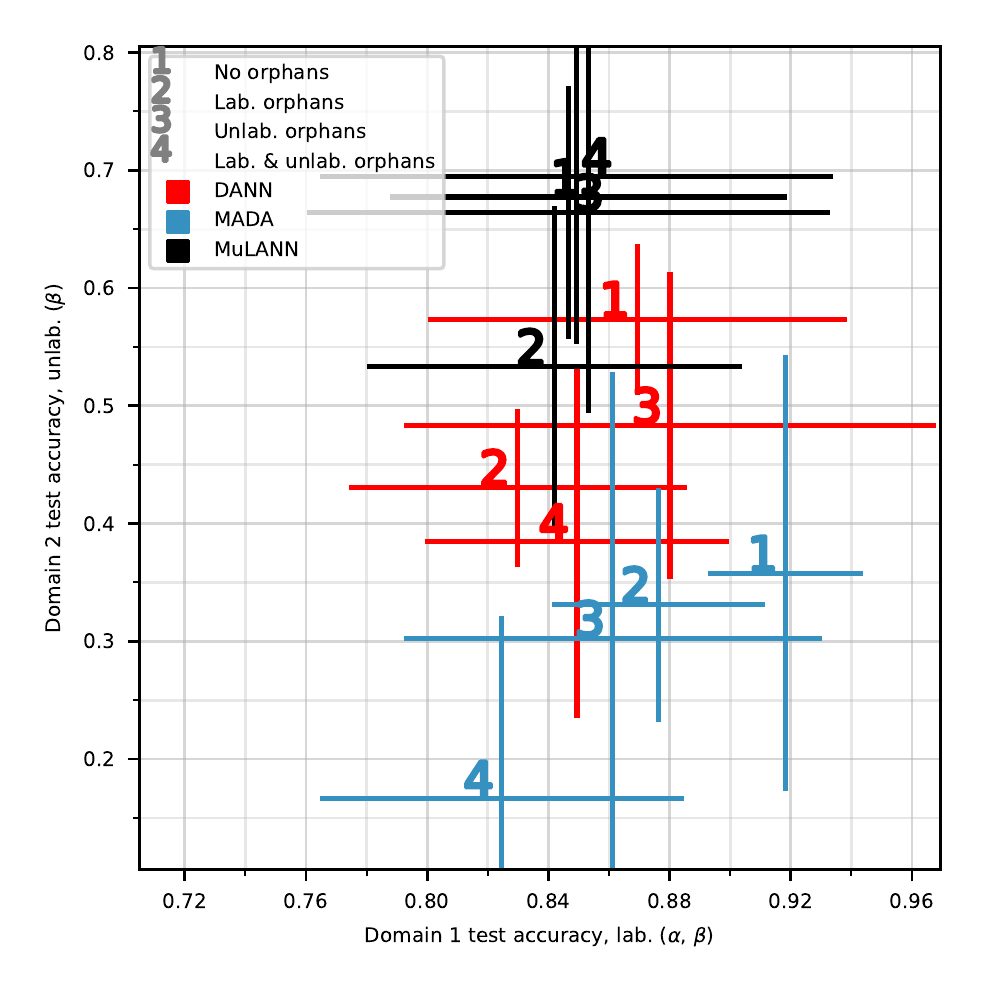}}
\caption{Impact of asymmetry in class content between domains on \pheno\ (T$\leftrightarrow$E) for \DANN, MADA and \XX.}
\label{fig:asymmetry_cell}
\end{figure}

\end{document}